\documentclass[11pt]{article}

\usepackage[preprint]{acl}
\usepackage{times}
\usepackage{latexsym}
\usepackage[T1]{fontenc}
\usepackage[utf8]{inputenc}
\usepackage{microtype}
\usepackage{inconsolata}
\usepackage{graphicx}

\usepackage{booktabs}
\usepackage{array}
\usepackage{multirow}
\usepackage{amsmath}
\usepackage{amssymb}
\usepackage{pifont}
\usepackage[table]{xcolor}
\usepackage{soul}
\usepackage{algorithm}
\usepackage[noend]{algpseudocode}
\usepackage[most]{tcolorbox}
\usepackage{listings}
\tcbuselibrary{listings}
\usepackage{tikz}
\usepackage{float}
\usetikzlibrary{arrows.meta,positioning,fit,backgrounds,calc}

\definecolor{GoogleBlue}{RGB}{66,133,244}
\definecolor{GoogleGreen}{RGB}{52,168,83}
\definecolor{GoogleRed}{RGB}{234,67,53}
\definecolor{GooglePurple}{RGB}{123,31,162}
\definecolor{GoogleOrange}{RGB}{255,153,0}
\definecolor{GoogleTeal}{RGB}{0,150,136}
\definecolor{GoogleBrown}{RGB}{121,85,72}
\colorlet{refinebg}{red!7}

\newcommand{\styleprism}{\textsc{StylePrism}}
\newcommand{\sprefine}{\textsc{SP-Refine}}
\newcommand{\spinfer}{\textsc{SP-Infer}}
\newcommand{\spprofile}{\textsc{SP-Profile}}
\newcommand{\spregen}{\textsc{SP-Regen}}
\newcommand{\spregenp}{\textsc{SP-Regen+P}}
\newcommand{\oregen}{\textsc{Oracle-Regen}}
\newcommand{\oregenp}{\textsc{Oracle-Regen+P}}
\newcommand{\osel}{\textsc{Oracle-Select}}
\newcommand{\randuser}{\textsc{Random-Samples}}
\newcommand{\noprof}{\textsc{No Profile}}
\newcommand{\longlamp}{\textsc{LongLaMP}}   
\newcommand{\sfid}{SF}
\newcommand{\tq}{TQ}

\colorlet{L}{GoogleBlue}\colorlet{S}{GoogleGreen}\colorlet{D}{GoogleRed}\colorlet{O}{GoogleOrange}
\colorlet{P}{GooglePurple}\colorlet{F}{GoogleTeal}\colorlet{X}{GoogleBrown}
\colorlet{Lbg}{GoogleBlue!17}\colorlet{Sbg}{GoogleGreen!20}\colorlet{Dbg}{GoogleRed!15}\colorlet{Obg}{GoogleOrange!22}
\colorlet{Pbg}{GooglePurple!13}\colorlet{Fbg}{GoogleTeal!17}\colorlet{Xbg}{GoogleBrown!17}
\newcommand{\an}[3]{{\sethlcolor{#1bg}\hl{#3}}\textsuperscript{\textcolor{#1}{\bfseries\tiny #2}}}
\newcommand{\dv}[3]{{\setulcolor{#1}\setul{0.3ex}{0.13ex}\ul{#3}}\textsuperscript{\textcolor{#1}{\bfseries\tiny $\neg$#2}}}
\newcommand{\mv}[2]{{\tiny\textcolor{#1!85!black}{\bfseries[#2]}}}
\newcommand{\pc}[2]{\textcolor{#1}{\bfseries #2}\,}

\tcbset{promptbox/.style={colback=gray!5, colframe=gray!60, boxrule=0.4pt,
  arc=1mm, left=1.5mm, right=1.5mm, top=1mm, bottom=1mm, fonttitle=\bfseries\small}}
\tcbset{qualbox/.style={colback=white, colframe=gray!55, boxrule=0.4pt, arc=0.8mm,
  left=1.4mm, right=1.4mm, top=1mm, bottom=1mm, fonttitle=\bfseries\scriptsize, fontupper=\scriptsize}}

\algrenewcommand\algorithmicrequire{\textbf{Input:}}
\algrenewcommand\algorithmicensure{\textbf{Output:}}

\title{Explicit Style Extraction for Personalized Long-Form Text Generation}
\author{
  Jenitha Reddy Patel\textsuperscript{1} \quad
  Sameeksha Kini\textsuperscript{1} \quad
  Amulya Rayasam\textsuperscript{1} \quad
  Anisa Dye\textsuperscript{1} \quad
  Abhinav Sharma\textsuperscript{2} \\
  \bfseries Sai Karthik Navuluru\textsuperscript{3} \quad
  Bo Ni\textsuperscript{4} \quad
  Li Li\textsuperscript{5} \quad
  Junda Wu\textsuperscript{6,7} \quad
  Tong Yu\textsuperscript{6} \quad
  Nedim Lipka\textsuperscript{6} \\
  \bfseries Xiaolei Huang\textsuperscript{8,9} \quad
  Tyler Derr\textsuperscript{4} \quad
  Yue Zhao\textsuperscript{5} \quad
  Zhengzhong Tu\textsuperscript{10} \quad
  Yu Wang\textsuperscript{11,12} \quad
  Hongjie Chen\textsuperscript{13} \\
  \bfseries Thien Huu Nguyen\textsuperscript{12,14} \quad
  Nesreen K. Ahmed\textsuperscript{15} \quad
  Franck Dernoncourt\textsuperscript{6} \quad
  Ryan A. Rossi\textsuperscript{16} \\[4pt]
  \normalfont\normalsize
  \textsuperscript{1}University of Chicago \quad
  \textsuperscript{2}University of Massachusetts Amherst \quad
  \textsuperscript{3}University of Texas at Dallas \\
  \textsuperscript{4}Vanderbilt University \quad
  \textsuperscript{5}University of Southern California \quad
  \textsuperscript{6}Adobe Research \quad
  \textsuperscript{7}UCSD \\
  \textsuperscript{8}University of Missouri \quad
  \textsuperscript{9}University of Memphis \quad
  \textsuperscript{10}Texas A\&M University \quad
  \textsuperscript{11}University of Georgia \\
  \textsuperscript{12}University of Oregon \quad
  \textsuperscript{13}Dolby Laboratories \quad
  \textsuperscript{14}Amazon \quad
  \textsuperscript{15}Cisco \quad
  \textsuperscript{16}Zillow \\[2pt]
  \texttt{jenithapatel@uchicago.edu}
}

\begin{document}
\maketitle

\begin{abstract}
Personalized long-form generation asks a language model to write new text the way a specific user would. The standard recipe, retrieval-augmented generation, places a few of the user's past texts in the prompt but leaves the model to decide what to take from them. We introduce \styleprism{}, a taxonomy of 28 stylistic aspects in seven categories, and \sprefine{}, an inference-time method that makes style extraction an explicit step: the model first fills \styleprism{} with the user's patterns, quoted evidence, and habits to avoid, then drafts from the user's samples and revises the draft twice against this profile, guided by its own style scores. On the three long-form tasks of LongLaMP with three LLMs, \sprefine{} outperforms LongLaMP's retrieval-augmented baseline in 25 of 27 ROUGE and METEOR comparisons, by 10.2\% on average and by up to 35.8\%. Annotating users' own texts with the extracted profiles shows that \styleprism{} captures concrete, checkable habits, from the order of rhetorical moves to a reviewer's sign-off. Ablations show that the profile helps as revision guidance, not as a replacement for the samples: writing from the profile alone is worse in all 27 comparisons, while a single implicit-inference call is a cheaper alternative that comes close. Oracle variants that let the reviser see the reference, together with reference-free controls, show that larger refinement gains require information unavailable at inference time.
\end{abstract}

\section{Introduction}

Large language models (LLMs) are increasingly used to draft text on a person's behalf: research abstracts, product reviews, social media posts. For such text to be useful, it should be correct and fluent and also sound like its author. Personalized text generation studies this problem by conditioning a model on a user's writing history \citep{salemi-etal-2024-lamp, zhang2024personalization}, and the LongLaMP benchmark \citep{kumar2024longlamp} extends it to long-form outputs, where differences in style between users are most visible.

The dominant approach is retrieval-augmented personalization: retrieve the user's past texts that are most relevant to the current input and place them in the prompt \citep{salemi-etal-2024-lamp, kumar2024longlamp, salemi2024optimization}. Retrieval decides \emph{which} samples the model sees, but says little about \emph{what} the model should take from them. In LongLaMP, the retrieved samples are followed by a single instruction such as ``Following the given patterns''. Whether the model attends to sentence rhythm, hedging, discourse moves, or formatting is left implicit.

We make style an explicit, inspectable intermediate. \styleprism{} (Figure~\ref{fig:styleprism}) organizes writing style into 28 aspects in seven categories, from lexical choice to cross-task fingerprints. \sprefine{} (Algorithm~\ref{alg:refine}) uses it as a function: given the retrieved samples, the model first \emph{extracts} a style profile that fills every \styleprism{} aspect with an observed pattern, quoted evidence, and habits to avoid; it then drafts the output from the samples, and revises the draft twice against the profile, scoring its own revision on five style criteria and fixing the weakest. Unlike retrieval-augmented generation, \sprefine{} produces a profile that a person can read and check (Figure~\ref{fig:gt-abstract}).

\begin{figure*}[t]
\centering
\includegraphics[width=\textwidth]{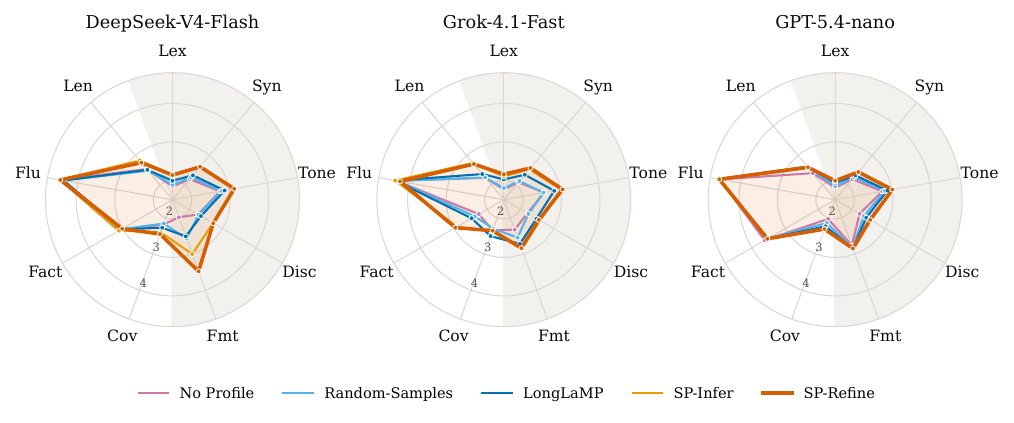}
\caption{LLM-judge scores (1--5) on the nine criteria, averaged over the three LongLaMP tasks, for each generator. The shaded half holds the five style criteria (Style Fidelity): Lexical (Lex), Syntax (Syn), Tone, Discourse (Disc), and Format (Fmt); the other half holds the four task-quality criteria: Coverage (Cov), Factuality (Fact), Fluency (Flu), and Length (Len). For all three generators, \sprefine{} (vermillion) and \spinfer{} (orange) form the outer boundary on the style criteria, ahead of \longlamp{} (tied with it only on Format for GPT). \sprefine{}'s gains over \longlamp{} are largest with DeepSeek, where Format rises by almost one point, and smallest with GPT. \noprof{} and \randuser{} are shown for reference. Per-criterion values are in Appendix~\ref{app:judge}.}
\label{fig:radar}
\end{figure*}

We evaluate on the three long-form LongLaMP tasks with three LLM families, using reference-based metrics and a rubric-based LLM judge that scores style fidelity and task quality separately (Figure~\ref{fig:radar}). Our contributions are:
\begin{itemize}
  \item \styleprism{}, a taxonomy of writing style that serves as the schema for extraction, the guide for revision, and the basis of our style rubric.
  \item \sprefine{}, which outperforms LongLaMP's retrieval-augmented baseline in 25 of 27 reference-based comparisons and improves LLM-judged style fidelity over it in 8 of 9 generator--task settings.
  \item An annotation study of users' reference texts showing which profile aspects are borne out by what the user actually wrote, and where users depart from their own profile.
  \item Ablations and an oracle study showing that the profile works as revision guidance but not as a replacement for the samples, and separating gains achievable at inference time from those that need the reference.
\end{itemize}

\begin{figure*}[t]
\centering
\scriptsize
\renewcommand{\arraystretch}{1.0}
\newcommand{\asp}[4]{$\bullet$\,\textcolor{#1}{\textbf{#2}}\,\textbf{#3} {\tiny -#4}\\}
\newcommand{\cat}[2]{\textbf{\textsc{\textcolor{#1}{#2}}}\\}
\begin{tabular}{@{}p{0.488\textwidth}@{\hspace{0.024\textwidth}}p{0.488\textwidth}@{}}
\toprule
\begin{tabular}[t]{@{}l@{}}
\cat{GoogleBlue}{Lexical Choice}
\asp{GoogleBlue}{L1}{Word Frequency \& Preference}{Detects idiosyncrasies and recurring words}
\asp{GoogleBlue}{L2}{Stopword Avoidance / Usage}{Does user avoid starting with ``The'', ``This'', etc.}
\asp{GoogleBlue}{L3}{Domain-Specific Lexicon}{Detects technical or field-specific vocabulary}
\asp{GoogleBlue}{L4}{Intensifier \& Modifier Use}{Detects degree of assertiveness or emphasis}
\asp{GoogleBlue}{L5}{Hedging Behavior}{Detects cautious vs.\ assertive phrasing patterns}
\midrule
\cat{GoogleGreen}{Syntactic Patterns}
\asp{GoogleGreen}{S1}{Sentence Length/Complexity}{Avg.\ sentence length and clause depth}
\asp{GoogleGreen}{S2}{Passive vs.\ Active Voice}{Ratio of passive to active constructions}
\asp{GoogleGreen}{S3}{Coordination vs.\ Subordination}{Preference for connecting or nesting ideas}
\asp{GoogleGreen}{S4}{Punctuation Habits}{Frequency of commas, parentheses, semicolons, etc.}
\asp{GoogleGreen}{S5}{Enumerative Style}{Frequency of lists or structured enumerations}
\midrule
\cat{GooglePurple}{Pragmatic \& Tone}
\asp{GooglePurple}{P1}{Formality Level}{Degree of formality or conversational tone.}
\asp{GooglePurple}{P2}{Politeness Strategy}{Softening vs.\ direct tone and mitigation strategies.}
\asp{GooglePurple}{P3}{Audience Awareness}{Inclusive or direct reference to reader.}
\asp{GooglePurple}{P4}{Figurative Language}{Use of analogy, metaphor, or idiom.}
\asp{GooglePurple}{P5}{Certainty \& Commitment}{Confidence conveyed via modal verbs or stance.}
\end{tabular}
&
\begin{tabular}[t]{@{}l@{}}
\cat{GoogleOrange}{Structural Organization}
\asp{GoogleOrange}{O1}{Paragraph Structure}{Average paragraph length and internal cohesion.}
\asp{GoogleOrange}{O2}{Rhetorical Role Distribution}{Freq.\ of problem, contribution, \& implication}
\asp{GoogleOrange}{O3}{Argument Framing}{Logical order and transition structure.}
\asp{GoogleOrange}{O4}{Section Naming Style}{Naming consistency and section organization.}
\midrule
\cat{GoogleRed}{Discourse Markers}
\asp{GoogleRed}{D1}{Transitional Phrases}{Recurrent connective or bridging expressions}
\asp{GoogleRed}{D2}{Opening Moves}{Typical ways of beginning paragraphs or sections}
\asp{GoogleRed}{D3}{Closing Moves}{How arguments or sections conclude}
\asp{GoogleRed}{D4}{Reference Density}{Frequency of citations or attributions}
\midrule
\cat{GoogleTeal}{Formatting \& Layout}
\asp{GoogleTeal}{F1}{Visual Structure}{Use of bullets, numbering, or headings.}
\asp{GoogleTeal}{F2}{Paragraph Density}{Consistency of idea segmentation.}
\asp{GoogleTeal}{F3}{Emphasis \& Capitalization}{Use of visual emphasis in text.}
\midrule
\cat{GoogleBrown}{Cross-Task Variability}
\asp{GoogleBrown}{X1}{Style Adaptation by Task}{How writing style changes across task types.}
\asp{GoogleBrown}{X2}{Content--Style Entanglement}{Consistency of stylistic cues across content.}
\end{tabular}
\\
\bottomrule
\end{tabular}
\caption{\textbf{The \styleprism{} framework:} 28 stylistic aspects in seven categories that \sprefine{} extracts from a user's past writing and uses to revise new text. Codes (L1--X2) and colors are used throughout the paper; descriptions and examples are in Appendix~\ref{app:styleprism}.}
\label{fig:styleprism}
\end{figure*}

\section{Related Work}

\paragraph{Personalized text generation.}
LaMP \citep{salemi-etal-2024-lamp} introduced a benchmark for personalizing LLM outputs with retrieved user history, and LongLaMP \citep{kumar2024longlamp} extended it to long-form generation. Most methods follow the retrieval-augmented recipe and improve the retriever \citep{salemi2024optimization, mysore2023pearl}, summarize the user's history \citep{richardson2023integrating}, or decompose personalization into multiple stages \citep{li2023teach}; see \citet{zhang2024personalization} for a survey. We keep retrieval fixed and study what the model is told to extract from the retrieved samples and how it uses it.

\paragraph{Representing writing style.}
Style has been studied in style representation learning, where a central difficulty is separating style from content \citep{wegmann2022same}, and in style transfer \citep{jin2022tst}. LLMs can transfer style from instructions and a few examples \citep{reif2022recipe} and describe stylistic attributes in natural language \citep{patel2023lisa}. \styleprism{} organizes such attributes into a schema that an LLM fills from a handful of samples, with quoted evidence for each claim.

\paragraph{Refinement and LLM-based evaluation.}
LLMs can critique and revise their own outputs \citep{madaan2023selfrefine}, and LLM judges are widely used to evaluate generation \citep{zheng2023judging, liu2023geval}, although they can favor their own outputs \citep{panickssery2024selfpref}. \sprefine{}'s revisions are guided by an explicit style profile and never see the reference; we compare them with oracle revisions that do.

\section{Method}
\label{sec:method}

\subsection{Task Setup}
Each test instance consists of a user $u$ with a history $P_u = \{d_1, \dots, d_m\}$ of previously written documents, a task input $x$ (e.g., a paper title and key items) of task type $c$, and a reference output $y$ written by $u$. A generator LLM $M$ must produce $\hat{y}$ that is faithful to $x$ and written in $u$'s style. Following LongLaMP, all personalized methods first retrieve the $k$ entries of $P_u$ most relevant to $x$, $S = R(x, P_u, k)$. The retriever $R$ follows LongLaMP's construction: the query is the task input (for abstracts, only its list of key items); each history entry is represented by the concatenation of its fields (title and abstract; summary and post; rating, summary, product description, and review); query and entries are embedded with Contriever \citep{izacard2022contriever} by mean-pooling its token embeddings; and the $k$ entries with the highest dot-product score are returned. We use $k=4$, and all methods use the same $S$. The LongLaMP baseline then generates $\hat y = M(\phi_{\text{LL}}(S, x))$, where the template $\phi_{\text{LL}}$ joins the samples and appends the task input with an instruction such as ``Following the given patterns''.

\subsection{\styleprism{}}
\label{sec:styleprism}
\styleprism{} (Figure~\ref{fig:styleprism}) has seven categories: \emph{lexical choice} (L1--L5), \emph{syntactic patterns} (S1--S5), \emph{discourse markers} (D1--D4), \emph{structural organization} (O1--O4), \emph{pragmatic tone} (P1--P5), \emph{formatting and layout} (F1--F3), and \emph{cross-task variability} (X1--X2), which records how style shifts across task types and which fingerprints persist across content. We write $\Sigma$ for the set of 28 aspects; descriptions and examples are in Appendix~\ref{app:styleprism}.

\subsection{\sprefine{}}
\label{sec:sprefine}
\sprefine{} (Algorithm~\ref{alg:refine}; full prompts in Appendix~\ref{app:prompts}) has four steps, each one call to the generator $M$, and never sees the reference. Steps 1 and 2 are the extraction call of \spprofile{} and the single call of \spinfer{} (\S\ref{sec:ablations}); in our experiments their outputs are computed once per instance and reused, so the refinement stage itself (steps 3 and 4) adds two calls.

\paragraph{1. Extract style.} The prompt $\phi_{\text{ext}}$ casts $M$ as a computational linguist analyzing the $k$ samples of one author. It requires every claim to be based on the text, exact phrases to be quoted as evidence, no pattern to be inferred that is not present, and ``not observable from samples'' to be written otherwise. $M$ returns the profile $\pi$, a JSON object with one entry per aspect $\sigma \in \Sigma$. Each entry holds a natural-language \texttt{pattern}, a list of quoted \texttt{evidence}, and (for 22 aspects) an \texttt{avoid} field; some add typed fields such as average words per sentence, passive percentage, or Boolean use of bullets and headers (Appendix~\ref{app:psp}).

\paragraph{2. Draft.} The prompt $\phi_{\text{draft}}$ shows the numbered samples and the task, and asks $M$ to silently identify the author's recurring style features (sentence rhythm, word choices, hedging, transitions, paragraph structure, tone, formatting), draft the output, check the draft against the samples, and return only the final text, $y_0$. This step does not see $\pi$.

\paragraph{3. Revise and score.} The prompt $\phi_{\text{rev}}$ casts $M$ as a writing evaluator and ghostwriter. It receives $S$, $\pi$, a task reminder (``complete this task and only this task''), and $y_0$. $M$ first revises $y_0$ using $\pi$ and $S$ as style anchors, applying every applicable profile field (sentence length, voice, hedging, transitional phrases, paragraph density, formatting) without altering or adding facts, giving $y_1$. It then scores $y_1$ against the samples from 1 to 5 on five style criteria $A$ = \{Lexical, Syntactic, Tone \& Formality, Discourse \& Structure, Formatting\}, each with anchored extremes, and names its weakest criteria and specific issues; we write $a_1 \in \{1,\dots,5\}^5$ for the scores.

\paragraph{4. Targeted revision.} The prompt $\phi_{\text{tgt}}$ receives $S$, $\pi$, the task reminder, $y_1$, and $a_1$ with the named issues. $M$ revises $y_1$ in a fixed order of priority: criteria scored 1--2 first, then criteria scored 3, leaving criteria scored 4--5 unchanged, again without altering or adding facts. It rescores the result; these scores are not used. The output is $\hat y = y_2$.

\begin{algorithm}[t]
\caption{\sprefine{}. Inputs: task input $x$ of type $c$; user history $P_u$; retriever $R$ and number of samples $k$; generator LLM $M$; \styleprism{} schema $\Sigma$ (28 aspects). $\phi_{\cdot}$ are prompt templates (Appendix~\ref{app:prompts}); $M(\cdot)$ is one LLM call. $\pi$ maps each aspect $\sigma$ to its pattern, quoted evidence $E_\sigma$, and avoid field; $A$ is the set of five style criteria; $a_t \in \{1,\dots,5\}^{|A|}$ are $M$'s scores of draft $y_t$ against $S$. Steps 1 and 2 coincide with \spprofile{}'s extraction and \spinfer{}; their outputs are reused, so the refinement stage adds two calls (steps 3--4).}
\label{alg:refine}
\footnotesize
\begin{algorithmic}[1]
\Require $x, c, P_u, R, k, M, \Sigma$
\Ensure personalized output $\hat y$
\State $S \gets R(x, P_u, k)$ \Comment{retrieve $k$ samples}
\State $\pi \gets \textsc{ExtractStyle}(S, \Sigma)$ \Comment{step 1}
\State $y_0 \gets M(\phi_{\text{draft}}(S, c, x))$ \Comment{step 2}
\State $(y_1, a_1) \gets \textsc{Revise}(S, \pi, c, x, y_0)$ \Comment{step 3}
\State $y_2 \gets \textsc{TargetedRevise}(S, \pi, c, x, y_1, a_1)$ \Comment{step 4}
\State \Return $\hat y \gets y_2$
\Statex
\Function{ExtractStyle}{$S, \Sigma$}
  \State \Return $\pi = \{\sigma \mapsto (\text{pattern}_\sigma, E_\sigma, \text{avoid}_\sigma)\}_{\sigma \in \Sigma} \gets M(\phi_{\text{ext}}(S, \Sigma))$
\EndFunction
\Function{Revise}{$S, \pi, c, x, y_0$} \Comment{one call, $\phi_{\text{rev}}$}
  \State $y_1 \gets y_0$ rewritten to match every applicable field of $\pi$, anchored on $S$; facts of $x$ unchanged
  \State $a_1 \gets$ scores of $y_1$ against $S$ on each criterion in $A$, with the weakest criteria and specific issues
  \State \Return $(y_1, a_1)$
\EndFunction
\Function{TargetedRevise}{$S, \pi, c, x, y_1, a_1$} \Comment{one call, $\phi_{\text{tgt}}$}
  \State fix $\{j : a_{1,j} \le 2\}$, then improve $\{j : a_{1,j} = 3\}$; keep $\{j : a_{1,j} \ge 4\}$
  \State \Return revised draft $y_2$ \Comment{rescores; scores unused}
\EndFunction
\end{algorithmic}
\end{algorithm}

\section{Experimental Setup}

\paragraph{Data.}
We use the three public long-form tasks of LongLaMP \citep{kumar2024longlamp} in the \emph{user} setting, in which test users do not appear in training: abstract generation (title and key items $\rightarrow$ abstract), review writing (rating, product description, and summary $\rightarrow$ review), and topic writing (Reddit post summary $\rightarrow$ post). We evaluate on the full test sets: 4,560 abstract, 1,822 review, and 2,452 topic instances (Appendix~\ref{app:data}). No method is trained or tuned on these data.

\paragraph{Models.}
We generate with three LLMs from different providers, DeepSeek-V4-Flash, Grok-4.1-Fast (non-reasoning), and GPT-5.4-nano, accessed through Azure AI Foundry. Every call uses the same client and settings: the system message ``You are a helpful assistant.'' and the prompt as the user message; style extraction requests JSON output. A response blocked by the provider's content filter is kept as an empty output and scored as such.

\paragraph{Baselines.}
\emph{\noprof{}} sends the task input alone (e.g., ``Generate an abstract for the title \dots{} using the following items: \dots''), with no samples and no added instruction. \emph{\longlamp{}} formats the $k$ retrieved entries with LongLaMP's own prompt templates \citep{kumar2024longlamp} and appends the task input, e.g., after ``Following the given patterns'' for reviews; the entries' text is passed through unchanged. \emph{\randuser{}} uses the same template, but with $k$ entries drawn at random from the user's own history instead of the $k$ most relevant ones, which isolates the effect of retrieval. As in LongLaMP, profile abstracts are truncated to their first 750 words in all methods. All prompts are in Appendix~\ref{app:prompts}.

\paragraph{Reference-based metrics.}
As LongLaMP does, we report ROUGE-1, ROUGE-L \citep{lin2004rouge}, and METEOR \citep{banerjee2005meteor}, computed with the LongLaMP evaluation code and averaged over test instances.

\paragraph{LLM judge.}
\label{sec:judge}
Reference-overlap metrics reward shared content but say little about style. We therefore also score outputs with an LLM judge, GPT-5.4-nano with low reasoning effort. The judge sees the task, the full generated output, and the full reference, and rates the output from 1 to 5 on nine criteria with anchored descriptions of the extremes. \emph{Style Fidelity} (\sfid{}) is the mean of Lexical, Syntax, Tone, Discourse, and Format; \emph{Task Quality} (\tq{}) is the mean of Coverage, Factuality, Fluency, and Length. Both use the reference as the anchor; \sprefine{}'s own scores in steps 3--4 use the same five style criteria but the samples as the anchor, so the method never sees the judge's anchor. Because the judge is also one of our generators, it may favor GPT-generated text \citep{panickssery2024selfpref}; we therefore compare methods only within a generator.

\section{Results}
\label{sec:results}

\begin{table*}[t]
\centering
\small
\setlength{\tabcolsep}{5pt}
\begin{tabular}{@{}l ccc ccc ccc@{}}
\toprule
 & \multicolumn{3}{c}{\textbf{Abstract Generation}} & \multicolumn{3}{c}{\textbf{Review Writing}} & \multicolumn{3}{c}{\textbf{Topic Writing}} \\
\cmidrule(lr){2-4}\cmidrule(lr){5-7}\cmidrule(lr){8-10}
\textbf{Method} & R-1 & R-L & MET & R-1 & R-L & MET & R-1 & R-L & MET \\
\midrule
\multicolumn{10}{@{}l}{\textit{DeepSeek-V4-Flash}} \\
\noprof{} & 0.3773 & 0.2026 & 0.2835 & 0.2910 & 0.1384 & 0.1685 & 0.2871 & 0.1256 & 0.2180 \\
\randuser{} & 0.3904 & 0.2073 & 0.3028 & 0.3229 & 0.1582 & 0.1983 & 0.2717 & 0.1294 & 0.1979 \\
\longlamp{} & 0.4185 & 0.2344 & 0.3352 & 0.3345 & 0.1704 & 0.2114 & 0.2818 & 0.1427 & 0.2077 \\
\rowcolor{refinebg}\sprefine{} & \textbf{0.4426} & \textbf{0.2639} & \textbf{0.3535} & \textbf{0.3872} & \textbf{0.1852} & \textbf{0.2806} & \textbf{0.3265} & \textbf{0.1604} & \textbf{0.2585} \\
\midrule
\multicolumn{10}{@{}l}{\textit{Grok-4.1-Fast}} \\
\noprof{} & 0.3469 & 0.1825 & 0.2714 & 0.3111 & 0.1305 & 0.2394 & 0.2625 & 0.1108 & 0.2208 \\
\randuser{} & 0.3502 & 0.1855 & 0.2720 & 0.3055 & 0.1272 & 0.2495 & 0.2641 & 0.1146 & 0.2095 \\
\longlamp{} & 0.3902 & 0.2221 & 0.3065 & 0.3444 & 0.1582 & \textbf{0.2900} & 0.3049 & 0.1486 & \textbf{0.2485} \\
\rowcolor{refinebg}\sprefine{} & \textbf{0.4318} & \textbf{0.2692} & \textbf{0.3273} & \textbf{0.3510} & \textbf{0.1684} & 0.2744 & \textbf{0.3052} & \textbf{0.1528} & 0.2360 \\
\midrule
\multicolumn{10}{@{}l}{\textit{GPT-5.4-nano}} \\
\noprof{} & 0.3539 & 0.1908 & 0.2511 & 0.2487 & 0.1267 & 0.1313 & 0.2758 & 0.1206 & 0.2028 \\
\randuser{} & 0.3667 & 0.1979 & 0.2704 & 0.3113 & 0.1425 & 0.1827 & 0.2755 & 0.1211 & 0.2129 \\
\longlamp{} & 0.3857 & 0.2126 & 0.2902 & 0.3171 & 0.1463 & 0.1885 & 0.2854 & 0.1280 & 0.2183 \\
\rowcolor{refinebg}\sprefine{} & \textbf{0.4027} & \textbf{0.2280} & \textbf{0.3411} & \textbf{0.3489} & \textbf{0.1529} & \textbf{0.2559} & \textbf{0.3048} & \textbf{0.1415} & \textbf{0.2322} \\
\bottomrule
\end{tabular}
\caption{Results on the LongLaMP user setting (full test sets). R-1/R-L: ROUGE-1/ROUGE-L; MET: METEOR. Best per column within each generator in \textbf{bold}. Relative gains over \longlamp{} are in Table~\ref{tab:gains} and Figure~\ref{fig:bars} (Appendix~\ref{app:more}); ablations are in Table~\ref{tab:ablation}.}
\label{tab:main}
\end{table*}

\begin{table}[t]
\centering
\footnotesize
\setlength{\tabcolsep}{3.2pt}
\begin{tabular}{@{}l cc cc cc@{}}
\toprule
 & \multicolumn{2}{c}{\textbf{Abstract}} & \multicolumn{2}{c}{\textbf{Review}} & \multicolumn{2}{c}{\textbf{Topic}} \\
\cmidrule(lr){2-3}\cmidrule(lr){4-5}\cmidrule(lr){6-7}
\textbf{Method} & \sfid{} & \tq{} & \sfid{} & \tq{} & \sfid{} & \tq{} \\
\midrule
\multicolumn{7}{@{}l}{\textit{DeepSeek-V4-Flash}} \\
\noprof{} & 2.80 & 3.42 & 2.00 & 3.09 & 1.82 & 2.72 \\
\randuser{} & 3.04 & 3.36 & 2.24 & 3.10 & 1.81 & 2.57 \\
\longlamp{} & 3.12 & 3.42 & 2.26 & 3.13 & 1.86 & 2.60 \\
\rowcolor{refinebg}\sprefine{} & \textbf{3.49} & \textbf{3.53} & \textbf{2.69} & \textbf{3.17} & \textbf{2.28} & \textbf{2.81} \\
\midrule
\multicolumn{7}{@{}l}{\textit{Grok-4.1-Fast}} \\
\noprof{} & 2.99 & 3.11 & 1.80 & 2.75 & 1.78 & 2.52 \\
\randuser{} & 3.08 & 3.21 & 1.81 & 2.68 & 1.88 & 2.54 \\
\longlamp{} & 3.27 & 3.36 & 2.07 & 2.75 & 2.10 & 2.67 \\
\rowcolor{refinebg}\sprefine{} & \textbf{3.40} & \textbf{3.46} & \textbf{2.32} & \textbf{3.03} & \textbf{2.18} & \textbf{2.70} \\
\midrule
\multicolumn{7}{@{}l}{\textit{GPT-5.4-nano}} \\
\noprof{} & 3.10 & 3.58 & 2.15 & 3.04 & 1.80 & 2.82 \\
\randuser{} & 3.16 & 3.62 & 2.29 & 3.23 & 1.82 & 2.74 \\
\longlamp{} & 3.24 & \textbf{3.67} & \textbf{2.34} & \textbf{3.25} & 1.87 & 2.80 \\
\rowcolor{refinebg}\sprefine{} & \textbf{3.39} & 3.59 & 2.18 & \textbf{3.25} & \textbf{2.13} & \textbf{2.92} \\
\bottomrule
\end{tabular}
\caption{LLM-judge scores (GPT-5.4-nano, 1--5). \sfid{}: Style Fidelity; \tq{}: Task Quality. Best per column within each generator in \textbf{bold}. Full test sets. Per-criterion scores are in Appendix~\ref{app:judge}.}
\label{tab:judge}
\end{table}

\paragraph{Reference-based metrics.}
Table~\ref{tab:main} compares \sprefine{} with the baselines. \sprefine{} scores highest in 25 of 27 generator--task--metric comparisons. It is above \noprof{} in all 27, so the gains come from using the user's writing; \longlamp{} is above \randuser{} in all 27, so choosing the most relevant samples matters as well. Relative to \longlamp{}, which sees the same retrieved samples, \sprefine{} improves by 10.2\% on average (14.9\% with DeepSeek, 4.4\% with Grok, 11.5\% with GPT; Table~\ref{tab:gains}). The largest gains are on METEOR for review writing: $+$35.8\% with GPT (0.2559 vs.\ 0.1885) and $+$32.7\% with DeepSeek (0.2806 vs.\ 0.2114). The two exceptions are METEOR on review and topic writing with Grok, where \longlamp{} is higher (0.2900 vs.\ 0.2744 and 0.2485 vs.\ 0.2360).

\paragraph{LLM judge.}
Table~\ref{tab:judge} shows that \sprefine{} improves Style Fidelity over \longlamp{} in 8 of 9 generator--task settings, by 0.21 points on average, and over \noprof{} in all 9. Its Task Quality is higher than \longlamp{}'s in 7 settings, equal in 1, and lower in 1 (GPT abstracts, 3.59 vs.\ 3.67), so the style gains do not come at the expense of content. The exception on Style Fidelity is GPT on review writing (2.18 vs.\ 2.34), where Format scores are low for every method that sees the samples (\S\ref{sec:ablres}). Figure~\ref{fig:radar} breaks the scores down by criterion: with DeepSeek, \sprefine{}'s gain over \longlamp{} spans all five style criteria and is largest for Format ($+$0.97); with Grok, it is largest on Syntax and Tone; with GPT, all differences are within 0.13.

\section{What Does Style Extraction Capture?}
\label{sec:qual}

\begin{figure*}[t]
\begin{tcolorbox}[qualbox, title={Reference abstract written by the user (abstract generation, idx\_13; profile extracted by Grok-4.1-Fast from $k{=}4$ other abstracts)}]
\begin{minipage}[t]{0.655\linewidth}
\setlength{\parskip}{0pt}\linespread{1.08}\selectfont
\mv{GoogleOrange}{O2: problem} \an{L}{L4}{Accurate} and timely \an{L}{L1}{traffic classification} is a key to providing \an{S}{S5}{Quality of Service (QoS), application-level visibility, and security monitoring} for network operations and management. A class of traffic classification techniques have emerged \an{S}{S3}{that apply machine learning technology to predict the application class of a traffic flow based on the statistical properties of flow-features}. \mv{GoogleOrange}{O2: method} \an{D}{D1}{In this paper}, \dv{P}{P3}{we propose} a \an{L}{L4}{novel} iterative-tuning scheme to increase the training speed of the classification algorithm using \an{L}{L3}{Support Vector Machine (SVM)} learning. Meanwhile \dv{P}{P3}{we derive} the equations to obtain SVM parameters by conducting theoretical analysis of iterative-tuning SVM. \mv{GoogleOrange}{O2: evaluation} Traffic classification \an{S}{S2}{is carried out using} flow-level information extracted from \an{L}{L3}{NetFlow} data. \mv{GoogleOrange}{O2: results} Performance evaluation demonstrates that \an{X}{X2}{the proposed} iterative-tuning SVM exhibits a training speed that is \an{O}{O3}{two to ten times faster than eight other previously proposed SVM techniques} found in the literature, while maintaining comparable classification accuracy as those eight SVM techniques. \dv{D}{D3}{In the presence of millions of flows and Terabytes of data in the network, faster training speeds is essential to making SVM techniques a viable option for real-world deployment of traffic classification modules. In addition, network operators and cloud service providers can apply network traffic classification to address a range of issues including semi-real-time security monitoring and traffic engineering.}
\end{minipage}\hfill
\begin{minipage}[t]{0.325\linewidth}
\tiny\setlength{\parskip}{1.2pt}
\textbf{What the profile extracted} (abridged; full profile in Appendix~\ref{app:psp})\par
\pc{L}{L1} repeats key technical terms\par
\pc{L}{L3} networking lexicon: ``NetFlow v5'', ``HTTP tunnels''\par
\pc{L}{L4} sparse modifiers; prefers ``novel'', ``accurate''\par
\pc{S}{S1} long multi-clause sentences, 24 words on average\par
\pc{S}{S2} $\sim$40\% passive: ``is carried out using information''\par
\pc{S}{S3} subordination: ``using information that is only available\dots''\par
\pc{S}{S5} comma-separated series of three or more items\par
\pc{D}{D1} ``This paper presents'', ``In this paper''\par
\pc{D}{D3} closes on a performance claim: ``outperforms them by at least 10\%''\par
\pc{O}{O2} problem $\rightarrow$ method $\rightarrow$ evaluation $\rightarrow$ results\par
\pc{O}{O3} superiority shown by comparison and metrics\par
\pc{P}{P3} impersonal, no personal pronouns\par
\pc{F}{F1}\pc{F}{F2} one dense paragraph, no headers or lists\par
\pc{X}{X2} fingerprints: ``novel/heuristic proposal'', ``performance metrics at end''
\end{minipage}
\end{tcolorbox}
\vspace{-2mm}
\caption{What \styleprism{} extraction captures. Only the user's actual abstract is shown; each highlight marks text matching an aspect in the profile extracted \emph{before} this abstract was seen, colored by category (Figure~\ref{fig:styleprism}). Orange bracketed tags such as \mv{GoogleOrange}{O2: method} mark the rhetorical moves the profile predicts (O2). Colored underlines with $\neg$ mark where the user departs from the profile: first person (P3) and a closing on applications instead of a performance claim (D3). Whole-text matches: formal academic register (P1), one unbroken paragraph without headers (F1--F2), and 25.5 words per sentence (S1; profile: 24).}
\label{fig:gt-abstract}
\end{figure*}

Aggregate scores do not show whether the extracted profile describes the user. Because $\pi$ is built from the $k$ retrieved samples only, the user's reference text for the instance is held out from extraction and can serve as a check. Figure~\ref{fig:gt-abstract} annotates one reference abstract with the aspects of its profile (Grok-4.1-Fast); Figures~\ref{fig:gt-review} and~\ref{fig:gt-topic} (Appendix~\ref{app:annot}) do the same for a review and a Reddit post.

\paragraph{The profile predicts concrete, checkable habits.}
Extracted from four other abstracts, the profile predicts the abstract's rhetorical order (problem, method, evaluation, results; O2), its transition ``In this paper'' (D1), the qualifiers ``accurate'' and ``novel'' (L4), a passive evaluation sentence built on the exact phrase ``is carried out using'' (S2), a comma-separated series (S5), superiority shown by comparison (``two to ten times faster than eight other \dots{} techniques''; O3), the single unbroken paragraph (F1--F2), and the sentence length (25.5 words vs.\ 24 predicted; S1). For the reviewer, the profile recovers the sign-off ``[2008 hyperbolium dot com]'' verbatim as a fingerprint (X2), the contrast-based evaluation against earlier versions of each song (O3), and even a metaphor the profile quotes from an earlier review (``nosing their way''; the reference has ``nosed its way towards disco's rhythms'', P4).

\paragraph{Users depart from their own profile.}
The abstract's author writes in the first person (``we propose'', ``we derive'') although the profile, built from impersonal samples, says to avoid it (P3), and closes on applications rather than a performance claim (D3). The reviewer omits the fractional star rating the profile lists as a fingerprint, and the Reddit user gives suggestions inline rather than as a numbered list (S5). Style varies across a user's documents, which bounds how closely any profile-based method can match one reference.

\section{Ablations}
\label{sec:ablations}

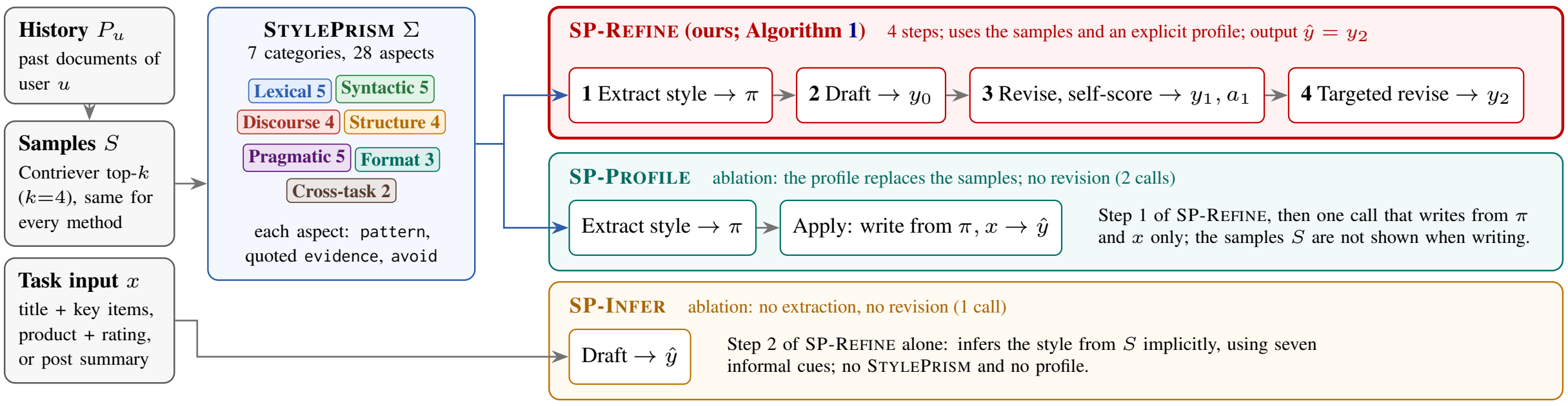
\begin{figure*}[t]
\centering
\resizebox{\textwidth}{!}{%
\begin{tikzpicture}[
  font=\scriptsize, >=Stealth,
  box/.style={draw=#1!70!black, fill=#1!6, rounded corners=1.2mm, line width=0.5pt, align=left, inner sep=1.6mm},
  step/.style={draw=#1!75!black, fill=white, rounded corners=0.8mm, line width=0.45pt, align=center, inner xsep=1.4mm, inner ysep=1mm, minimum height=6.2mm},
  chip/.style={fill=#1!14, draw=#1!80!black, rounded corners=0.5mm, line width=0.35pt, inner sep=0.6mm, font=\tiny\bfseries, text=#1!70!black},
  arr/.style={->, line width=0.55pt, draw=black!55},
  ttl/.style={anchor=north west, font=\scriptsize\bfseries, inner sep=0pt},
  desc/.style={anchor=west, font=\tiny, align=left, inner sep=0pt},
]
\node[box=gray, text width=1.6cm] (hist) at (0,1.45) {\textbf{History} $P_u$\\[1pt]{\tiny past documents of user $u$}};
\node[box=gray, text width=1.6cm] (ret) at (0,0.0) {\textbf{Samples} $S$\\[1pt]{\tiny Contriever top-$k$ ($k{=}4$), same for every method}};
\node[box=gray, text width=1.6cm] (task) at (0,-1.55) {\textbf{Task input} $x$\\[1pt]{\tiny title + key items, product + rating, or post summary}};
\draw[arr] (hist) -- (ret);
\node[box=GoogleBlue, text width=2.7cm, align=center] (sp) at (2.85,0.45) {
  \textbf{\styleprism{}} $\Sigma$\\[-1pt]{\tiny 7 categories, 28 aspects}\\[3pt]
  \tikz\node[chip=GoogleBlue]{Lexical 5};\,\tikz\node[chip=GoogleGreen]{Syntactic 5};\\[2pt]
  \tikz\node[chip=GoogleRed]{Discourse 4};\,\tikz\node[chip=GoogleOrange]{Structure 4};\\[2pt]
  \tikz\node[chip=GooglePurple]{Pragmatic 5};\,\tikz\node[chip=GoogleTeal]{Format 3};\\[2pt]
  \tikz\node[chip=GoogleBrown]{Cross-task 2};\\[3pt]
  {\tiny each aspect: \texttt{pattern}, quoted \texttt{evidence}, \texttt{avoid}}};
\def\lx{5.2}\def\pad{0.22}
\node[ttl, text=red!70!black] (rt) at (\lx+\pad,1.83) {\sprefine{} (ours; Algorithm~\ref{alg:refine})\quad\normalfont\tiny 4 steps; uses the samples and an explicit profile; output $\hat y = y_2$};
\node[step=red, anchor=west] (r1) at (\lx+\pad,1.0) {\textbf{1} Extract style $\rightarrow \pi$};
\node[step=red, right=2.6mm of r1] (r2) {\textbf{2} Draft $\rightarrow y_0$};
\node[step=red, right=2.6mm of r2] (r3) {\textbf{3} Revise, self-score $\rightarrow y_1, a_1$};
\node[step=red, right=2.6mm of r3] (r4) {\textbf{4} Targeted revise $\rightarrow y_2$};
\draw[arr] (r1) -- (r2); \draw[arr] (r2) -- (r3); \draw[arr] (r3) -- (r4);
\node[ttl, text=GoogleTeal!70!black] (pt) at (\lx+\pad,0.18) {\spprofile{}\quad\normalfont\tiny ablation: the profile replaces the samples; no revision (2 calls)};
\node[step=GoogleTeal, anchor=west] (p1) at (\lx+\pad,-0.5) {Extract style $\rightarrow \pi$};
\node[step=GoogleTeal, right=2.6mm of p1] (p2) {Apply: write from $\pi, x$ $\rightarrow \hat y$};
\draw[arr] (p1) -- (p2);
\node[desc, text width=5.0cm] (pd) at ($(p2.east)+(4mm,0)$) {Step 1 of \sprefine{}, then one call that writes from $\pi$ and $x$ only; the samples $S$ are not shown when writing.};
\node[ttl, text=GoogleOrange!65!black] (it) at (\lx+\pad,-1.28) {\spinfer{}\quad\normalfont\tiny ablation: no extraction, no revision (1 call)};
\node[step=GoogleOrange, anchor=west] (i1) at (\lx+\pad,-1.96) {Draft $\rightarrow \hat y$};
\node[desc, text width=7.0cm] (id) at ($(i1.east)+(4mm,0)$) {Step 2 of \sprefine{} alone: infers the style from $S$ implicitly, using seven informal cues; no \styleprism{} and no profile.};
\node[fit=(rt)(r4)(pt)(pd)(it)(id), inner sep=0pt] (allc) {};
\begin{scope}[on background layer]
\node[fit=(rt)(r1)(r4), inner sep=0pt] (rf) {};
\node[fit=(pt)(p1)(pd), inner sep=0pt] (pf) {};
\node[fit=(it)(i1)(id), inner sep=0pt] (if) {};
\draw[draw=red!75!black, fill=red!5, rounded corners=1.2mm, line width=0.9pt] let \p1=(allc.east), \p2=(rf.south), \p3=(rf.north) in
  (\lx cm,\y2-1.6mm) rectangle (\x1+2.4mm,\y3+1.6mm);
\draw[draw=GoogleTeal!75!black, fill=GoogleTeal!5, rounded corners=1.2mm, line width=0.5pt] let \p1=(allc.east), \p2=(pf.south), \p3=(pf.north) in
  (\lx cm,\y2-1.6mm) rectangle (\x1+2.4mm,\y3+1.6mm);
\draw[draw=GoogleOrange!75!black, fill=GoogleOrange!5, rounded corners=1.2mm, line width=0.5pt] let \p1=(allc.east), \p2=(if.south), \p3=(if.north) in
  (\lx cm,\y2-1.6mm) rectangle (\x1+2.4mm,\y3+1.6mm);
\end{scope}
\draw[arr] (ret.east) -- ($(sp.west)+(0,-0.45)$);
\coordinate (bus) at ($(sp.east)+(0.32,0)$);
\draw[arr, draw=GoogleBlue!70!black] (sp.east) -- (bus) |- (r1.west);
\draw[arr, draw=GoogleBlue!70!black] (bus) |- (p1.west);
\draw[arr] (task.east) -- ++(0.25,0) |- (i1.west);
\end{tikzpicture}}
\caption{\sprefine{} and its ablations. $M$ is the generator LLM; all methods receive the same retrieved samples $S$ and task input $x$ (and the task type $c$, omitted). \sprefine{} makes style extraction an explicit function: step~1 fills the \styleprism{} schema $\Sigma$ with the user's patterns, quoted evidence, and things to avoid, giving the profile $\pi$; step~2 drafts $y_0$ from $S$; step~3 revises the draft against $\pi$ and $S$ and scores the revision $y_1$ on five style criteria ($a_1$); step~4 fixes the lowest-scored criteria, giving $y_2$. \spprofile{} keeps only step~1 and then writes from $\pi$ without $S$; \spinfer{} keeps only step~2.}
\label{fig:methods}
\end{figure*}

\begin{table*}[t]
\centering
\scriptsize
\setlength{\tabcolsep}{3.4pt}
\begin{tabular}{@{}l cccc cccc cccc@{}}
\toprule
 & \multicolumn{4}{c}{\textbf{Abstract Generation}} & \multicolumn{4}{c}{\textbf{Review Writing}} & \multicolumn{4}{c}{\textbf{Topic Writing}} \\
\cmidrule(lr){2-5}\cmidrule(lr){6-9}\cmidrule(lr){10-13}
\textbf{Variant} & R-L & MET & \sfid{} & \tq{} & R-L & MET & \sfid{} & \tq{} & R-L & MET & \sfid{} & \tq{} \\
\midrule
\multicolumn{13}{@{}l}{\textit{DeepSeek-V4-Flash}} \\
\rowcolor{refinebg}\sprefine{} & 0.264 & 0.353 & \textbf{3.49} & 3.53 & 0.185 & 0.281 & \textbf{2.69} & 3.17 & 0.160 & \textbf{0.259} & 2.28 & 2.81 \\
\spinfer{} & \textbf{0.273} & \textbf{0.360} & 3.40 & \textbf{3.56} & \textbf{0.186} & \textbf{0.291} & 2.47 & 3.18 & \textbf{0.164} & 0.258 & \textbf{2.32} & \textbf{2.86} \\
\spprofile{} & 0.247 & 0.325 & 3.46 & 3.51 & 0.165 & 0.225 & 2.61 & \textbf{3.20} & 0.143 & 0.219 & 2.23 & 2.82 \\
\midrule
\multicolumn{13}{@{}l}{\textit{Grok-4.1-Fast}} \\
\rowcolor{refinebg}\sprefine{} & \textbf{0.269} & \textbf{0.327} & \textbf{3.40} & 3.46 & \textbf{0.168} & 0.274 & 2.32 & 3.03 & \textbf{0.153} & \textbf{0.236} & 2.18 & 2.70 \\
\spinfer{} & 0.254 & 0.320 & \textbf{3.40} & \textbf{3.51} & 0.167 & \textbf{0.278} & 2.27 & \textbf{3.04} & 0.149 & 0.236 & \textbf{2.19} & \textbf{2.75} \\
\spprofile{} & 0.218 & 0.297 & 3.31 & 3.36 & 0.145 & 0.219 & \textbf{2.34} & 3.03 & 0.127 & 0.221 & 1.99 & 2.58 \\
\midrule
\multicolumn{13}{@{}l}{\textit{GPT-5.4-nano}} \\
\rowcolor{refinebg}\sprefine{} & \textbf{0.228} & \textbf{0.341} & \textbf{3.39} & 3.59 & \textbf{0.153} & \textbf{0.256} & 2.18 & 3.25 & \textbf{0.141} & \textbf{0.232} & 2.13 & 2.92 \\
\spinfer{} & 0.217 & 0.332 & 3.36 & \textbf{3.60} & 0.151 & 0.253 & 2.17 & 3.24 & 0.139 & 0.229 & \textbf{2.14} & \textbf{2.94} \\
\spprofile{} & 0.215 & 0.313 & \textbf{3.39} & 3.59 & 0.149 & 0.210 & \textbf{2.48} & \textbf{3.28} & 0.130 & 0.217 & \textbf{2.14} & 2.86 \\
\bottomrule
\end{tabular}
\caption{\sprefine{} and its ablations (full test sets). \spinfer{} keeps only the drafting call; \spprofile{} keeps the extraction call but writes from the profile without the samples. Best per column within each generator in \textbf{bold}. ROUGE-1 for all variants is in Figure~\ref{fig:bars}.}
\label{tab:ablation}
\end{table*}

\subsection{Variants}
Each ablation removes part of \sprefine{} (Figure~\ref{fig:methods}).

\paragraph{\spinfer{}: implicit inference, no extraction or revision.} \spinfer{} is a single call that receives the numbered samples and the task and is asked to identify the author's style silently, draft, check the draft against the samples, and return only the text (Appendix~\ref{app:prompts}). Its prompt names seven informal cues but does not use \styleprism{}, and no profile is produced. \sprefine{} uses this same call to produce its first draft, so comparing the two measures what explicit extraction and profile-guided revision add.

\paragraph{\spprofile{}: the profile replaces the samples.} \spprofile{} is step 1 of \sprefine{} followed by an \emph{apply} call that receives only $\pi$ and $x$. In one call, the apply prompt asks $M$ to draft while conforming to every applicable profile field, then to revise silently against it (e.g., sentence length within $\pm$20\% of the profile's value, every \texttt{avoid} item absent), without inventing facts. It tests whether a profile can stand in for the samples.

\subsection{Results}
\label{sec:ablres}

\paragraph{The profile helps as guidance, not as a replacement.}
\spprofile{} uses the same samples and the same profile as \sprefine{} yet scores below it in all 27 reference-based comparisons, and below \spinfer{} in all 27 (Table~\ref{tab:ablation}, Figure~\ref{fig:bars}). The per-criterion judge scores (Appendix~\ref{app:judge}) locate the loss: \spprofile{} is below \spinfer{} on Lexical in all 9 generator--task settings and on Coverage in 8 of 9, but above it on Format in 6 of 9, by 1.0--1.5 points on review writing. Writing from $\pi$ alone loses the samples' concrete wording and content cues, while explicit formatting fields are enforced reliably. \sprefine{} keeps the samples in view at every step and uses $\pi$ only to correct a draft, and it keeps \spprofile{}'s formatting gain on DeepSeek reviews (Format 3.37 vs.\ 3.32) while avoiding its lexical loss (1.90 vs.\ 1.83).

\paragraph{Implicit inference is a strong single call.}
Since \spinfer{} is \sprefine{}'s own first draft, the difference between the two is exactly what extraction and revision add. On ROUGE and METEOR this is small and depends on the generator: revision helps with Grok and GPT, where \sprefine{} is higher in 13 of 18 comparisons by up to 0.015, but hurts with DeepSeek, where it is lower in 8 of 9 by up to 0.010. On judged Style Fidelity, \sprefine{} is higher in 5 of 9 settings, tied in 1, and lower in 3 by at most 0.04. The largest gain is on DeepSeek reviews (2.69 vs.\ 2.47), almost entirely from Format. In short, \spinfer{} recovers most of \sprefine{}'s quality on its own, and the extraction and revision steps pay off mainly where the draft breaks the user's formatting conventions.

Revision does not repair every formatting problem. On review writing, the Format scores of Grok and GPT stay low after revision (1.84 and 1.91), well below \spprofile{}'s (2.84 and 3.36).

\paragraph{Why is \spinfer{} better than \sprefine{} with DeepSeek?}
A likely explanation is that, with DeepSeek, revision trades overlap with the reference for conformance to the profile. On abstracts and reviews, \sprefine{} raises DeepSeek's Style Fidelity over \spinfer{} (by 0.09 and 0.22), mostly through Format ($+$0.43 and $+$1.06), and raises Coverage on abstracts (3.23 vs.\ 3.13), yet its ROUGE and METEOR are lower in 8 of 9 comparisons. DeepSeek's drafts already have the highest overlap of the three generators (e.g., ROUGE-L 0.273 on abstracts vs.\ 0.254 for Grok and 0.217 for GPT). A revision that rewrites sentences toward the profile's patterns and evidence phrases, which come from \emph{other} documents, is then more likely to replace words the draft shared with the reference than to add new ones; with the weaker drafts of Grok and GPT, the same revisions repair more than they replace. DeepSeek may also follow ``apply every applicable PSP field'' more literally and rewrite more of the draft.

\subsection{Oracles and Reference-Free Controls}
\label{sec:oracle}
To bound what refinement can achieve, we evaluate variants that use information unavailable at inference time (Table~\ref{tab:ext}, Appendix~\ref{app:oracles}). \oregen{} gives a reviser the samples, the \spinfer{} output, and the reference; in one call it scores the output on nine criteria (style against the samples; coverage and length against the reference) and writes a revision, told to match the reference's length but not to copy its phrases. \oregenp{} starts from the \spprofile{} output instead. \osel{} keeps, per instance, the candidate that the reference-anchored judge scores highest among \{\spinfer{}, \oregen{}\}, among the three drafts \sprefine{} produces ($y_0$, $y_1$, $y_2$), or among \{\spprofile{}, \oregenp{}\}. The reference-free control \spregen{} uses the \oregen{} prompt with the reference block, the length target, and the copy instruction removed (Algorithm~\ref{alg:oracle}).

\paragraph{The reference, not the revision, drives the oracle gains.}
With the reference visible, \oregen{} raises ROUGE-L over \spinfer{} by 0.07--0.13 for DeepSeek and Grok but at most 0.016 for GPT, and \oregenp{} reaches ROUGE-L 0.490 on DeepSeek abstracts. Removing the reference (\spregen{}, Table~\ref{tab:control}) leaves the same revision with a ROUGE-L change of at most 0.014 and lower METEOR in 6 of 9 settings. The oracle outputs reuse the reference: for DeepSeek, 57\% of \oregen{} abstracts share an 8-word span with the reference that appears in neither the input nor the user's samples, compared with 13\% for \spinfer{}.

\paragraph{\sprefine{}'s intermediate drafts are not much better.}
Choosing the best of $y_0$, $y_1$, and $y_2$ with the reference-anchored judge (\osel{}(\sprefine{})) changes ROUGE-L over \sprefine{} by at most 0.008 for DeepSeek and 0.006 for Grok and GPT. The returned draft is close to the best that \sprefine{} produces; larger gains require new information, such as the reference.

\section{Conclusion}
We introduced \styleprism{}, a taxonomy of 28 stylistic aspects, and \sprefine{}, which uses it to make style extraction an explicit step in personalized long-form generation: the model writes down the user's style with quoted evidence, then drafts from the user's samples and revises against that profile. \sprefine{} outperforms LongLaMP's retrieval-augmented baseline in 25 of 27 reference-based comparisons and improves judged style fidelity, and the extracted profiles predict concrete habits that users' held-out texts bear out. Ablations show that the profile is useful as revision guidance but cannot replace the samples, and that a single implicit-inference call comes close. Oracle and control experiments show that larger refinement gains require the reference itself. Next steps are removing sample-format artifacts from the prompts and deciding per user or generator whether to revise.

\section*{Limitations}
We evaluate only the user setting of LongLaMP and only English data. Our \longlamp{} baseline uses LongLaMP's prompt templates and retrieval but our generators and generation settings, so absolute numbers are not directly comparable to those of \citet{kumar2024longlamp}. ROUGE and METEOR measure overlap with a single reference and capture style only indirectly. Our LLM judge is a model, not a human, and also one of our generators; its scores use the reference as an anchor, so Style Fidelity partly reflects content similarity. \sprefine{}'s self-scores use the same five style criteria as the judge, anchored on the samples rather than the reference, which may favor \sprefine{} on the judge's style criteria. The annotation study examines individual instances and is illustrative. We study prompting with a fixed retriever and $k=4$; other retrievers, values of $k$, and fine-tuning are not explored. Oracle controls use a sample of test instances, and closed-model APIs may change over time.

\section*{Ethics Statement}
Systems that imitate a person's writing style can be misused to impersonate that person or to produce deceptive text attributed to them. Our methods operate only on a user's own past writing for the purpose of assisting that user, and we do not release models trained to imitate specific individuals. Explicit profiles make what is being imitated inspectable, but they also summarize a person's habits, including signature phrases and sign-offs, and should be treated as personal data. Generated text can repeat sentences from a user's earlier documents and invent quantitative claims; deployed systems should check for both. We use the publicly released LongLaMP benchmark \citep{kumar2024longlamp} for research purposes only, consistent with its intended use; its Hugging Face release does not state a license. The data are derived from public scientific abstracts, Amazon reviews, and Reddit posts and may contain personal information, which we do not redistribute; the user texts shown in our figures are taken from the public benchmark.

\paragraph{Use of AI assistants.} We used AI assistants (Claude) to help write and debug experiment and analysis code, to format tables and figures, and to draft and edit parts of the text. The authors designed the study, ran all experiments, checked every number against the experiment outputs, and take full responsibility for the content.

\bibliography{references}

@inproceedings{salemi-etal-2024-lamp,
  title     = {{L}a{MP}: When Large Language Models Meet Personalization},
  author    = {Salemi, Alireza and Mysore, Sheshera and Bendersky, Michael and Zamani, Hamed},
  booktitle = {Proceedings of the 62nd Annual Meeting of the Association for Computational Linguistics (Volume 1: Long Papers)},
  pages     = {7370--7392},
  year      = {2024}
}

@article{kumar2024longlamp,
  title   = {Long{L}a{MP}: A Benchmark for Personalized Long-form Text Generation},
  author  = {Kumar, Ishita and Viswanathan, Snigdha and Yerra, Sushrita and Salemi, Alireza and Rossi, Ryan A. and Dernoncourt, Franck and Deilamsalehy, Hanieh and Chen, Xiang and Zhang, Ruiyi and Agarwal, Shubham and Lipka, Nedim and Van Nguyen, Chien and Nguyen, Thien Huu and Zamani, Hamed},
  journal = {arXiv preprint arXiv:2407.11016},
  year    = {2024}
}

@article{izacard2022contriever,
  title   = {Unsupervised Dense Information Retrieval with Contrastive Learning},
  author  = {Izacard, Gautier and Caron, Mathilde and Hosseini, Lucas and Riedel, Sebastian and Bojanowski, Piotr and Joulin, Armand and Grave, Edouard},
  journal = {Transactions on Machine Learning Research},
  year    = {2022}
}

@inproceedings{lin2004rouge,
  title     = {{ROUGE}: A Package for Automatic Evaluation of Summaries},
  author    = {Lin, Chin-Yew},
  booktitle = {Text Summarization Branches Out},
  pages     = {74--81},
  year      = {2004}
}

@inproceedings{banerjee2005meteor,
  title     = {{METEOR}: An Automatic Metric for {MT} Evaluation with Improved Correlation with Human Judgments},
  author    = {Banerjee, Satanjeev and Lavie, Alon},
  booktitle = {Proceedings of the {ACL} Workshop on Intrinsic and Extrinsic Evaluation Measures for Machine Translation and/or Summarization},
  pages     = {65--72},
  year      = {2005}
}

@inproceedings{salemi2024optimization,
  title     = {Optimization Methods for Personalizing Large Language Models through Retrieval Augmentation},
  author    = {Salemi, Alireza and Kallumadi, Surya and Zamani, Hamed},
  booktitle = {Proceedings of the 47th International ACM SIGIR Conference on Research and Development in Information Retrieval},
  year      = {2024}
}

@article{richardson2023integrating,
  title   = {Integrating Summarization and Retrieval for Enhanced Personalization via Large Language Models},
  author  = {Richardson, Chris and Zhang, Yao and Gillespie, Kellen and Kar, Sudipta and Singh, Arshdeep and Raeesy, Zeynab and Khan, Omar Zia and Sethy, Abhinav},
  journal = {arXiv preprint arXiv:2310.20081},
  year    = {2023}
}

@article{li2023teach,
  title   = {Teach {LLM}s to Personalize -- An Approach Inspired by Writing Education},
  author  = {Li, Cheng and Zhang, Mingyang and Mei, Qiaozhu and Wang, Yaqing and Hombaiah, Spurthi Amba and Liang, Yi and Bendersky, Michael},
  journal = {arXiv preprint arXiv:2308.07968},
  year    = {2023}
}

@inproceedings{mysore2023pearl,
  title     = {Pearl: Personalizing Large Language Model Writing Assistants with Generation-Calibrated Retrievers},
  author    = {Mysore, Sheshera and Lu, Zhuoran and Wan, Mengting and Yang, Longqi and Sarrafzadeh, Bahareh and Menezes, Steve and Baghaee, Tina and Gonzalez, Emmanuel Barajas and Neville, Jennifer and Safavi, Tara},
  booktitle = {Proceedings of the 1st Workshop on Customizable NLP: Progress and Challenges in Customizing NLP for a Domain, Application, Group, or Individual (CustomNLP4U)},
  year      = {2024}
}

@article{zhang2024personalization,
  title   = {Personalization of Large Language Models: A Survey},
  author  = {Zhang, Zhehao and Rossi, Ryan A. and Kveton, Branislav and Shao, Yijia and Yang, Diyi and Zamani, Hamed and Dernoncourt, Franck and others},
  journal = {arXiv preprint arXiv:2411.00027},
  year    = {2024}
}

@inproceedings{wegmann2022same,
  title     = {Same Author or Just Same Topic? Towards Content-Independent Style Representations},
  author    = {Wegmann, Anna and Schraagen, Marijn and Nguyen, Dong},
  booktitle = {Proceedings of the 7th Workshop on Representation Learning for NLP},
  pages     = {249--268},
  year      = {2022}
}

@inproceedings{patel2023lisa,
  title     = {Learning Interpretable Style Embeddings via Prompting {LLM}s},
  author    = {Patel, Ajay and Rao, Delip and Kothary, Ansh and McKeown, Kathleen and Callison-Burch, Chris},
  booktitle = {Findings of the Association for Computational Linguistics: EMNLP 2023},
  year      = {2023}
}

@article{jin2022tst,
  title   = {Deep Learning for Text Style Transfer: A Survey},
  author  = {Jin, Di and Jin, Zhijing and Hu, Zhiting and Vechtomova, Olga and Mihalcea, Rada},
  journal = {Computational Linguistics},
  volume  = {48},
  number  = {1},
  pages   = {155--205},
  year    = {2022}
}

@inproceedings{reif2022recipe,
  title     = {A Recipe for Arbitrary Text Style Transfer with Large Language Models},
  author    = {Reif, Emily and Ippolito, Daphne and Yuan, Ann and Coenen, Andy and Callison-Burch, Chris and Wei, Jason},
  booktitle = {Proceedings of the 60th Annual Meeting of the Association for Computational Linguistics (Volume 2: Short Papers)},
  pages     = {837--848},
  year      = {2022}
}

@inproceedings{madaan2023selfrefine,
  title     = {Self-Refine: Iterative Refinement with Self-Feedback},
  author    = {Madaan, Aman and Tandon, Niket and Gupta, Prakhar and Hallinan, Skyler and Gao, Luyu and Wiegreffe, Sarah and Alon, Uri and Dziri, Nouha and Prabhumoye, Shrimai and Yang, Yiming and Gupta, Shashank and Majumder, Bodhisattwa Prasad and Hermann, Katherine and Welleck, Sean and Yazdanbakhsh, Amir and Clark, Peter},
  booktitle = {Advances in Neural Information Processing Systems},
  year      = {2023}
}

@inproceedings{zheng2023judging,
  title     = {Judging {LLM}-as-a-Judge with {MT}-Bench and Chatbot Arena},
  author    = {Zheng, Lianmin and Chiang, Wei-Lin and Sheng, Ying and Zhuang, Siyuan and Wu, Zhanghao and Zhuang, Yonghao and Lin, Zi and Li, Zhuohan and Li, Dacheng and Xing, Eric P. and Zhang, Hao and Gonzalez, Joseph E. and Stoica, Ion},
  booktitle = {Advances in Neural Information Processing Systems (Datasets and Benchmarks Track)},
  year      = {2023}
}

@inproceedings{liu2023geval,
  title     = {{G}-Eval: {NLG} Evaluation using {GPT}-4 with Better Human Alignment},
  author    = {Liu, Yang and Iter, Dan and Xu, Yichong and Wang, Shuohang and Xu, Ruochen and Zhu, Chenguang},
  booktitle = {Proceedings of the 2023 Conference on Empirical Methods in Natural Language Processing},
  pages     = {2511--2522},
  year      = {2023}
}

@inproceedings{panickssery2024selfpref,
  title     = {{LLM} Evaluators Recognize and Favor Their Own Generations},
  author    = {Panickssery, Arjun and Bowman, Samuel R. and Feng, Shi},
  booktitle = {Advances in Neural Information Processing Systems},
  year      = {2024}
}

\appendix
\raggedbottom

\section{Dataset Statistics}
\label{app:data}
\begin{table}[H]
\centering
\footnotesize
\setlength{\tabcolsep}{3pt}
\begin{tabular}{@{}lrrrcc@{}}
\toprule
\textbf{Task} & \textbf{\#train} & \textbf{\#val} & \textbf{\#test} & \textbf{Input len.} & \textbf{Output len.} \\
\midrule
Abstract & 13,693 & 4,560 & 4,560 & 33.8$\pm$5.7 & 144.3$\pm$68.4 \\
Review   & 14,745 & 1,826 & 1,822 & 119.4$\pm$73.1 & 304.5$\pm$228.6 \\
Topic    & 11,442 & 2,452 & 2,452 & 28.4$\pm$36.1 & 263.0$\pm$243.3 \\
\bottomrule
\end{tabular}
\caption{LongLaMP user-setting statistics; lengths are mean $\pm$ s.d.\ in tokens as reported by \citet{kumar2024longlamp}. We evaluate on all test instances.}
\label{tab:data}
\end{table}

\section{\styleprism{} in Full}
\label{app:styleprism}
Table~\ref{tab:styleprism-full} describes each of the 28 aspects of \styleprism{} with an example. The codes used in our annotations (L1--X2) number the aspects in the order listed, as in Figure~\ref{fig:styleprism}.

\begin{table*}[t]
\centering
\caption{\textbf{Overview of our \styleprism{} framework.} Stylistic aspects that our approach learns from a set of text written by the user, which we then use for improving personalized text generation for that user.}
\label{tab:styleprism-full}
\scriptsize
\renewcommand{\arraystretch}{1.12}
\newcommand{\pcat}[2]{\textbf{\textsc{\textcolor{#1}{#2}}}}
\setlength{\tabcolsep}{4pt}
\begin{tabular}{@{}>{\raggedright\arraybackslash}p{2.15cm} >{\raggedright\arraybackslash}p{3.35cm} p{5.45cm} >{\raggedright\arraybackslash}p{4.0cm}@{}}
\toprule
\textbf{Category} & \textbf{Stylistic Aspect} & \textbf{Description} & \textbf{Example} \\
\midrule
\pcat{GoogleBlue}{Lexical Choice} & \textbf{Word Frequency \& Preference} & Detects recurring words, preferred synonyms, and idiosyncratic vocabulary. & Uses \textit{vitally important} instead of \textit{very important}. \\
 & \textbf{Stopword Avoidance / Usage} & Whether user avoids starting with ``The'', ``This'', etc. & Starts abstracts with content words. \\
 & \textbf{Domain-Specific Lexicon} & Detects technical or field-specific vocabulary tendencies. & Uses ``serves as'', ``gives rise to'', ``provides a foundation for''. \\
 & \textbf{Intensifier \& Modifier Use} & Detects degree of assertiveness or emphasis. & Frequent ``extremely'', ``remarkably'', ``fundamentally''. \\
 & \textbf{Hedging Behavior} & Detects cautious vs.\ assertive phrasing patterns. & ``May'', ``might'', ``could'' vs.\ ``is'', ``demonstrates''. \\
\midrule
\pcat{GoogleGreen}{Syntactic Patterns} & \textbf{Sentence Length \& Complexity} & Average sentence length, clause depth, punctuation frequency. & Long multi-clause vs.\ short direct sentences. \\
 & \textbf{Passive vs.\ Active Voice} & Ratio of passive to active constructions. & ``We propose'' vs.\ ``is proposed''. \\
 & \textbf{Coordination vs.\ Subordination} & Preference for connecting or nesting ideas. & ``Because'', ``although'' vs.\ ``and'', ``then''. \\
 & \textbf{Punctuation Habits} & Frequency of commas, parentheses, semicolons, etc. & Avoids dashes, prefers commas. \\
 & \textbf{Enumerative Style} & Frequency of lists or structured enumerations. & ``(1)\dots{} (2)\dots'' common in academic writing. \\
\midrule
\pcat{GoogleRed}{Discourse Markers} & \textbf{Transitional Phrases} & Recurrent connective or bridging expressions. & ``As a first step toward'', ``In practice'', ``The key challenge is that''. \\
 & \textbf{Opening Moves} & Typical ways of beginning paragraphs or sections. & ``We propose\dots'', ``This paper\dots'', or direct noun start. \\
 & \textbf{Closing Moves} & How arguments or sections conclude. & ``This provides a foundation\dots'', ``Overall, we demonstrate\dots''. \\
 & \textbf{Reference Density} & Frequency of citations or attributions. & Uses ``(Smith et al., 2023)'' often. \\
\midrule
\pcat{GoogleOrange}{Structural Organization} & \textbf{Paragraph Structure} & Average paragraph length and internal cohesion. & Short dense vs.\ long narrative paragraphs. \\
 & \textbf{Rhetorical Role Distribution} & Frequency of problem, contribution, and implication statements. & ``The goal is to\dots'', ``We make the following contributions\dots''. \\
 & \textbf{Argument Framing} & Logical order and transition structure. & ``The key challenge is that\dots{} We address this by\dots''. \\
 & \textbf{Section Naming Style} & Naming consistency and section organization. & Prefers ``Motivation'' over ``Introduction''. \\
\midrule
\pcat{GooglePurple}{Pragmatic \& Tone} & \textbf{Formality Level} & Degree of formality or conversational tone. & Emails: ``Hi'', ``Thanks'' vs.\ formal memos. \\
 & \textbf{Politeness Strategy} & Softening vs.\ direct tone and mitigation strategies. & ``I appreciate your input'' vs.\ ``This is incorrect''. \\
 & \textbf{Audience Awareness} & Inclusive or direct reference to reader. & ``We'' vs.\ ``you''. \\
 & \textbf{Figurative Language} & Use of analogy, metaphor, or idiom. & ``Gives rise to a new wave of methods.'' \\
 & \textbf{Certainty \& Commitment} & Confidence conveyed via modal verbs or stance. & ``Clearly shows'' vs.\ ``suggests''. \\
\midrule
\pcat{GoogleTeal}{Formatting \& Layout} & \textbf{Visual Structure} & Use of bullets, numbering, or headings. & Lists in emails or papers. \\
 & \textbf{Paragraph Density} & Consistency of idea segmentation. & One idea per paragraph vs.\ merged ideas. \\
 & \textbf{Emphasis \& Capitalization} & Use of visual emphasis in text. & ``IMPORTANT'', ``(!!!)'', etc. \\
\midrule
\pcat{GoogleBrown}{Cross-Task Variability} & \textbf{Style Adaptation by Task} & How writing style changes across task types. & Polite in casual emails to friends vs.\ professional emails. \\
 & \textbf{Content--Style Entanglement} & Consistency of stylistic cues across content. & ``Key challenge'' appears in papers and reviews. \\
\bottomrule
\end{tabular}
\end{table*}

\section{Additional Results}
\label{app:more}
\begin{figure}[t]
\centering
\includegraphics[width=\linewidth]{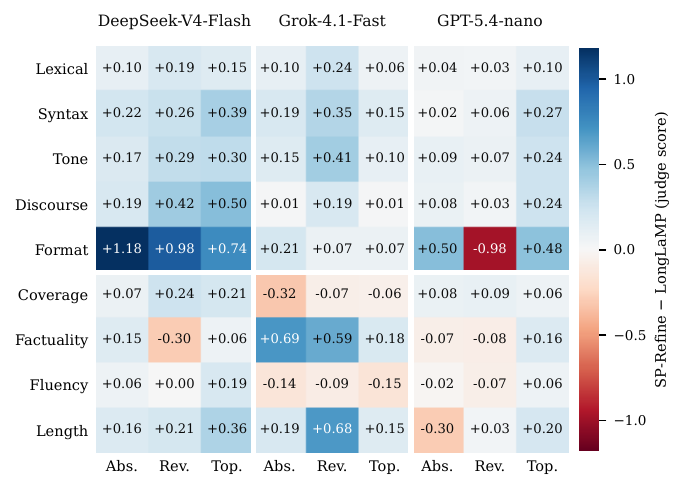}
\caption{Per-criterion LLM-judge difference between \sprefine{} and \longlamp{} (1--5 scale) for each generator and task (Abs.: abstract generation; Rev.: review writing; Top.: topic writing). Blue cells are gains for \sprefine{}, red cells losses. The upper block holds the five style criteria, the lower block the four task-quality criteria.}
\label{fig:heatmap}
\end{figure}

\begin{figure*}[t]
\centering
\includegraphics[width=\textwidth]{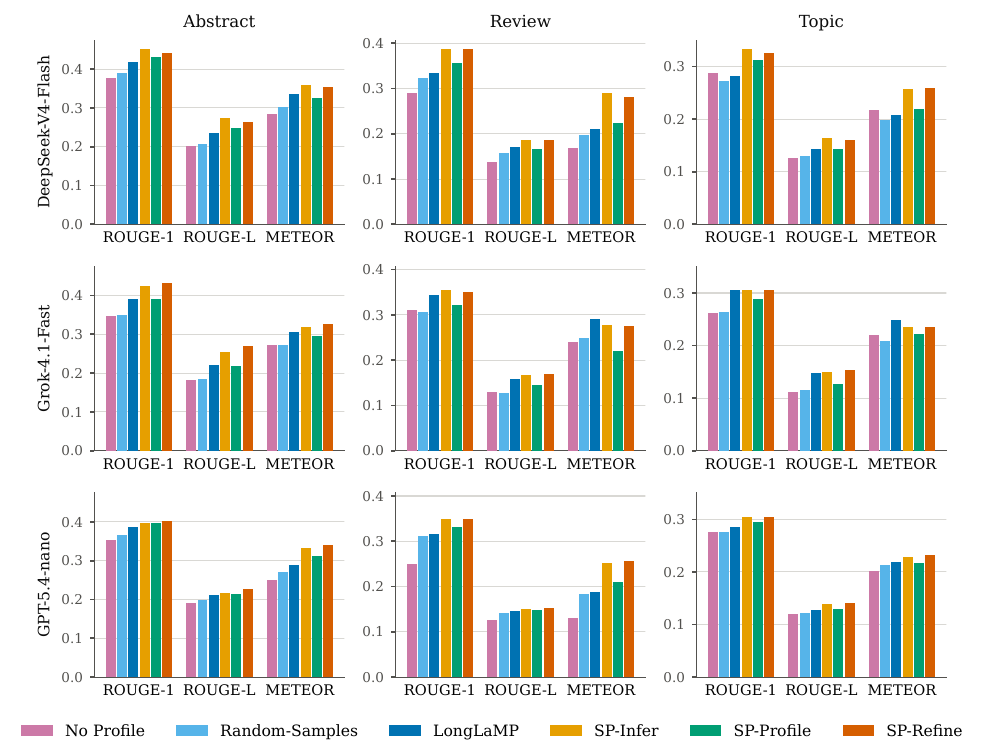}
\caption{ROUGE-1, ROUGE-L, and METEOR for \noprof{}, \longlamp{}, and the three \styleprism{} variants on each task and generator (full test sets; values in Tables~\ref{tab:main} and~\ref{tab:ablation}).}
\label{fig:bars}
\end{figure*}

Figure~\ref{fig:bars} plots ROUGE-1, ROUGE-L, and METEOR for the baselines and all \styleprism{} variants, and Table~\ref{tab:gains} gives the gains of \sprefine{} over \longlamp{} for every comparison. Figure~\ref{fig:heatmap} shows, for each judge criterion and generator--task setting, the difference between \sprefine{} and \longlamp{}: the style gains are largest for DeepSeek, mostly in Format, and the one loss in Style Fidelity (GPT on reviews) comes entirely from Format.

\begin{table*}[t]
\centering
\small
\setlength{\tabcolsep}{4.5pt}
\begin{tabular}{@{}l ccc ccc ccc@{}}
\toprule
 & \multicolumn{3}{c}{\textbf{Abstract Generation}} & \multicolumn{3}{c}{\textbf{Review Writing}} & \multicolumn{3}{c}{\textbf{Topic Writing}} \\
\cmidrule(lr){2-4}\cmidrule(lr){5-7}\cmidrule(lr){8-10}
 & R-1 & R-L & MET & R-1 & R-L & MET & R-1 & R-L & MET \\
\midrule
\multicolumn{10}{@{}l}{\textit{DeepSeek-V4-Flash}} \\
\longlamp{} & 0.4185 & 0.2344 & 0.3352 & 0.3345 & 0.1704 & 0.2114 & 0.2818 & 0.1427 & 0.2077 \\
\rowcolor{refinebg}\sprefine{} & 0.4426 & 0.2639 & 0.3535 & 0.3872 & 0.1852 & 0.2806 & 0.3265 & 0.1604 & 0.2585 \\
$\Delta$ & +0.0241 & +0.0295 & +0.0183 & +0.0527 & +0.0148 & +0.0692 & +0.0447 & +0.0177 & +0.0508 \\
$\Delta$\% & \textbf{+5.8} & \textbf{+12.6} & \textbf{+5.5} & \textbf{+15.8} & \textbf{+8.7} & \textbf{+32.7} & \textbf{+15.9} & \textbf{+12.4} & \textbf{+24.5} \\
\midrule
\multicolumn{10}{@{}l}{\textit{Grok-4.1-Fast}} \\
\longlamp{} & 0.3902 & 0.2221 & 0.3065 & 0.3444 & 0.1582 & 0.2900 & 0.3049 & 0.1486 & 0.2485 \\
\rowcolor{refinebg}\sprefine{} & 0.4318 & 0.2692 & 0.3273 & 0.3510 & 0.1684 & 0.2744 & 0.3052 & 0.1528 & 0.2360 \\
$\Delta$ & +0.0416 & +0.0471 & +0.0208 & +0.0066 & +0.0102 & $-$0.0156 & +0.0003 & +0.0042 & $-$0.0125 \\
$\Delta$\% & \textbf{+10.7} & \textbf{+21.2} & \textbf{+6.8} & \textbf{+1.9} & \textbf{+6.4} & $-$5.4 & \textbf{+0.1} & \textbf{+2.8} & $-$5.0 \\
\midrule
\multicolumn{10}{@{}l}{\textit{GPT-5.4-nano}} \\
\longlamp{} & 0.3857 & 0.2126 & 0.2902 & 0.3171 & 0.1463 & 0.1885 & 0.2854 & 0.1280 & 0.2183 \\
\rowcolor{refinebg}\sprefine{} & 0.4027 & 0.2280 & 0.3411 & 0.3489 & 0.1529 & 0.2559 & 0.3048 & 0.1415 & 0.2322 \\
$\Delta$ & +0.0170 & +0.0154 & +0.0509 & +0.0318 & +0.0066 & +0.0674 & +0.0194 & +0.0135 & +0.0139 \\
$\Delta$\% & \textbf{+4.4} & \textbf{+7.2} & \textbf{+17.5} & \textbf{+10.0} & \textbf{+4.5} & \textbf{+35.8} & \textbf{+6.8} & \textbf{+10.5} & \textbf{+6.4} \\
\bottomrule
\end{tabular}
\caption{Gains of \sprefine{} over the \longlamp{} baseline: absolute difference ($\Delta$) and relative difference ($\Delta$\%, positive values in bold). Averaged over all 27 comparisons, the relative gain is 10.2\%.}
\label{tab:gains}
\end{table*}

\section{Annotated Review and Topic Examples}
\label{app:annot}
\begin{figure*}[t]
\begin{tcolorbox}[qualbox, title={Reference review written by the user (review writing, idx\_0; profile extracted by Grok-4.1-Fast)}]
\begin{minipage}[t]{0.70\linewidth}
\setlength{\parskip}{0pt}\linespread{1.08}\selectfont
\an{D}{D2}{Soul singer Margie Joseph's third album for Atlantic, originally released in 1975,} revisited the \an{L}{L1}{funkier bottom end} of her 1973 label debut (confusingly titled Margie Joseph). \an{O}{O3}{Where 1974's Sweet Surrender had smoothed out the R\&B grooves} and \an{P}{P4}{nosed its way towards disco's rhythms}, this new set offered more grit on the upbeat numbers and deeper soul on the ballads. Arif Mardin continued in the producer's chair, and the song list was again constructed with a large dose of covers and a pair of originals. Joseph and Mardin's ``Sign of the Times'' opens the album with P Funk-styled synthesizer and plucked bass, and a melody that anticipates a seamless segue into Carole King's ``Believe in Humanity.'' Both songs offer up early-70s social spirituality in their lyrics. A cover of King's ``After All This Time'' appears later on the album, transitioning smoothly from a softer mid-tempo to a beefier soul shout as the strings and percussion start and stop for emphasis. ``The Same Love That Made Me Laugh'' is taken at a slow tempo, drawn out and \an{O}{O3}{more worn down than Bill Withers' original}, while ``Who Gets Your Love,'' is given a \an{O}{O3}{lighter treatment than Dusty Springfield's earlier version}. Joseph gets a chance to scat briefly at the end of ``Promise Me Your Love,'' unfortunately just in time for the fade, and the album closes with a cover of ``I Can't Move No Mountains'' that's decidedly more urban than Blood, Sweat \& Tears earlier horn-lined release. \an{D}{D3}{Many of Joseph's fans vote this as their favorite of her releases.} It's more sophisticated and less brash than her label debut, which is a plus or minus depending on \an{P}{P3}{your preference} for polish over roots. What this album makes clear is that Joseph was maturing as an artist, and her pairing with Mardin, which ended with this album, had developed over the course of their three albums together. \an{X}{X2}{[2008 hyperbolium dot com]}
\end{minipage}\hfill
\begin{minipage}[t]{0.27\linewidth}
\tiny\setlength{\parskip}{1.6pt}
\textbf{From the extracted profile}\par
\pc{D}{D2} opens with album history\par
\pc{L}{L1} precise style descriptors (``funky bottom end'')\par
\pc{O}{O3} contrasts with earlier albums and versions\par
\pc{P}{P4} metaphor: ``strings nosing their way into a few arrangements''\par
\pc{D}{D3} closes with an audience-targeted verdict\par
\pc{P}{P3} addresses the reader as ``you''\par
\pc{X}{X2} sign-off ``[2008 hyperbolium dot com]''
\end{minipage}
\end{tcolorbox}
\vspace{-2mm}
\caption{Review instance, annotated as in Figure~\ref{fig:gt-abstract} (only the main matches are marked). The profile, extracted from four other reviews, predicts the opening on album history (D2), contrasts with earlier versions (O3), an audience-targeted close (D3), and the reviewer's sign-off verbatim (X2); it even quotes a metaphor (``nosing their way'', P4) that the user reuses here.}
\label{fig:gt-review}
\end{figure*}
\begin{figure*}[t]
\begin{tcolorbox}[qualbox, title={Reference post written by the user (topic writing, idx\_1; profile extracted by Grok-4.1-Fast)}]
\begin{minipage}[t]{0.655\linewidth}
\setlength{\parskip}{2pt}\linespread{1.08}\selectfont
\an{D}{D2}{First off, I LOVE this game.} \an{D}{D1}{But} I'm only getting average \an{L}{L3}{arena wins} now \an{S}{S4}{(last ones were 5,4,6)}. I've done all the standard \an{L}{L1}{quests} and bought a few \an{L}{L1}{card packs} but after all that you kinda hit a wall.\par
If you don't pay the \$1.99 entry fee, you only get to use the \an{L}{L3}{Arena} every 2-3 days \an{S}{S4}{(assuming you do the daily quests)}. \an{S}{S3}{Or} you can win 45 games with one class \an{S}{S4}{(3 games @ 10 gold for 150 gold)}. \an{L}{L5}{I know} there are some special few who consistently get 7+ wins and are reimbursed entry, but thats not the average player.\par
Now in the past I used to have time and I grinded \an{L}{L3}{F2p} games like League of Legends. They felt pretty fair. Sure buying made it much faster, but you didn't feel like you were at a disadvantage. In Hearthstone, you need more then \an{L}{L3}{Basic cards} to do well in \an{L}{L3}{ranked} but grinding just doesn't make sense. Daily \an{L}{L1}{gold} is capped at Quest + 100g \an{S}{S4}{(if you grind for 30 wins)}. \an{L}{L3}{Crafting} doesn't make since either because you need cards to \an{L}{L3}{disenchant} or arena entries to collect \an{L}{L1}{dust}.\par
\an{D}{D1}{Since this is a BETA}, I would like to see is a somewhat more balanced system that increases gold output. \an{L}{L5}{I would suggest} \dv{S}{S5}{either allowing 2-3 daily quests or giving a random card (not pack) or dust for x wins.}
\end{minipage}\hfill
\begin{minipage}[t]{0.325\linewidth}
\tiny\setlength{\parskip}{1.2pt}
\textbf{What the profile extracted} (abridged)\par
\pc{L}{L1} game terms: ``gold'', ``Arena'', ``quests'', ``dust'', ``cards''\par
\pc{L}{L3} Hearthstone lexicon: ``F2P model'', ``daily quests'', ``Arena ticket''\par
\pc{L}{L5} occasional hedges: ``I believe'', ``I think''\par
\pc{S}{S1} mixes short and long sentences: ``Pay or leave.''\par
\pc{S}{S2} overwhelmingly active voice\par
\pc{S}{S3} coordination with ``and'', ``but'', ``or''\par
\pc{S}{S4} heavy use of parentheses\par
\pc{S}{S5} numbered lists for suggestions\par
\pc{D}{D1} ``But\dots'', ``Also'', ``Lastly''; evidence ``But\dots{} this is also a BETA''\par
\pc{D}{D2} starts with a summary or context\par
\textcolor{gray}{(remaining aspects omitted)}
\end{minipage}
\end{tcolorbox}
\vspace{-2mm}
\caption{Topic instance annotated as in Figure~\ref{fig:gt-abstract}. Lexical and syntactic aspects (game vocabulary, parenthetical asides, coordination) match closely; the user departs from the profile by giving suggestions inline instead of as a numbered list (S5).}
\label{fig:gt-topic}
\end{figure*}
Figures~\ref{fig:gt-review} and~\ref{fig:gt-topic} use the same markup as Figure~\ref{fig:gt-abstract}: highlights mark text that matches an aspect of the profile extracted from the user's retrieved samples, colored by \styleprism{} category; underlines with $\neg$ mark departures from the profile.

\section{Oracle Variants and Controls}
\label{app:oracles}
Table~\ref{tab:ext} reports every oracle variant on the full test sets (\S\ref{sec:oracle}); Table~\ref{tab:control} compares \oregen{} with its reference-free control \spregen{} on the same sample of instances.

\begin{table*}[t]
\centering
\footnotesize
\setlength{\tabcolsep}{2.6pt}
\begin{tabular}{@{}l ccc ccc ccc@{}}
\toprule
 & \multicolumn{3}{c}{\textbf{Abstract Generation}} & \multicolumn{3}{c}{\textbf{Review Writing}} & \multicolumn{3}{c}{\textbf{Topic Writing}} \\
\cmidrule(lr){2-4}\cmidrule(lr){5-7}\cmidrule(lr){8-10}
\textbf{Method} & R-1 & R-L & MET & R-1 & R-L & MET & R-1 & R-L & MET \\
\midrule
\multicolumn{10}{@{}l}{\textit{DeepSeek-V4-Flash}} \\
\spinfer{} & 0.452 & 0.273 & 0.360 & 0.387 & 0.186 & 0.291 & 0.334 & 0.164 & 0.258 \\
\rowcolor{refinebg}\sprefine{} & 0.443 & 0.264 & 0.353 & 0.387 & 0.185 & 0.281 & 0.327 & 0.160 & 0.259 \\
\midrule[0.2pt]
\rowcolor{gray!12}\oregen{}$^{\dagger}$ & \textit{0.549} & \textit{0.400} & \textit{0.484} & \textit{0.460} & \textit{0.261} & \textit{0.352} & \textit{0.398} & \textit{0.232} & \textit{0.323} \\
\rowcolor{gray!12}\oregenp{}$^{\dagger}$ & \textit{\textbf{0.620}} & \textit{\textbf{0.490}} & \textit{\textbf{0.548}} & \textit{\textbf{0.501}} & \textit{\textbf{0.305}} & \textit{\textbf{0.401}} & \textit{\textbf{0.429}} & \textit{\textbf{0.251}} & \textit{\textbf{0.351}} \\
\rowcolor{gray!12}\osel{}(\oregen{})$^{\dagger}$ & \textit{0.544} & \textit{0.393} & \textit{0.476} & \textit{0.459} & \textit{0.261} & \textit{0.355} & \textit{0.399} & \textit{0.233} & \textit{0.324} \\
\rowcolor{gray!12}\osel{}(\sprefine{})$^{\dagger}$ & \textit{0.451} & \textit{0.272} & \textit{0.360} & \textit{0.391} & \textit{0.189} & \textit{0.288} & \textit{0.334} & \textit{0.165} & \textit{0.261} \\
\rowcolor{gray!12}\osel{}(\oregenp{})$^{\dagger}$ & \textit{0.557} & \textit{0.407} & \textit{0.470} & \textit{0.449} & \textit{0.256} & \textit{0.336} & \textit{0.400} & \textit{0.222} & \textit{0.316} \\
\midrule
\multicolumn{10}{@{}l}{\textit{Grok-4.1-Fast}} \\
\spinfer{} & 0.424 & 0.254 & 0.320 & 0.355 & 0.167 & 0.278 & 0.306 & 0.149 & 0.236 \\
\rowcolor{refinebg}\sprefine{} & 0.432 & 0.269 & 0.327 & 0.351 & 0.168 & 0.274 & 0.305 & 0.153 & 0.236 \\
\midrule[0.2pt]
\rowcolor{gray!12}\oregen{}$^{\dagger}$ & \textit{\textbf{0.529}} & \textit{0.375} & \textit{\textbf{0.418}} & \textit{\textbf{0.438}} & \textit{\textbf{0.242}} & \textit{\textbf{0.355}} & \textit{\textbf{0.426}} & \textit{\textbf{0.268}} & \textit{\textbf{0.335}} \\
\rowcolor{gray!12}\oregenp{}$^{\dagger}$ & \textit{0.522} & \textit{\textbf{0.380}} & \textit{0.395} & \textit{0.423} & \textit{0.233} & \textit{0.303} & \textit{0.418} & \textit{0.264} & \textit{0.321} \\
\rowcolor{gray!12}\osel{}(\oregen{})$^{\dagger}$ & \textit{0.507} & \textit{0.347} & \textit{0.397} & \textit{0.431} & \textit{0.235} & \textit{0.349} & \textit{0.406} & \textit{0.244} & \textit{0.318} \\
\rowcolor{gray!12}\osel{}(\sprefine{})$^{\dagger}$ & \textit{0.430} & \textit{0.264} & \textit{0.325} & \textit{0.356} & \textit{0.169} & \textit{0.277} & \textit{0.306} & \textit{0.150} & \textit{0.233} \\
\rowcolor{gray!12}\osel{}(\oregenp{})$^{\dagger}$ & \textit{0.494} & \textit{0.341} & \textit{0.374} & \textit{0.410} & \textit{0.218} & \textit{0.290} & \textit{0.399} & \textit{0.240} & \textit{0.305} \\
\midrule
\multicolumn{10}{@{}l}{\textit{GPT-5.4-nano}} \\
\spinfer{} & 0.396 & 0.217 & 0.332 & \textbf{0.350} & 0.151 & 0.253 & 0.305 & 0.139 & 0.229 \\
\rowcolor{refinebg}\sprefine{} & 0.403 & 0.228 & \textbf{0.341} & 0.349 & 0.153 & \textbf{0.256} & 0.305 & 0.141 & \textbf{0.232} \\
\midrule[0.2pt]
\rowcolor{gray!12}\oregen{}$^{\dagger}$ & \textit{\textbf{0.411}} & \textit{\textbf{0.233}} & \textit{0.331} & \textit{0.348} & \textit{\textbf{0.154}} & \textit{0.234} & \textit{0.309} & \textit{\textbf{0.143}} & \textit{0.227} \\
\rowcolor{gray!12}\oregenp{}$^{\dagger}$ & \textit{0.408} & \textit{0.231} & \textit{0.313} & \textit{0.334} & \textit{0.149} & \textit{0.211} & \textit{0.301} & \textit{0.134} & \textit{0.215} \\
\rowcolor{gray!12}\osel{}(\oregen{})$^{\dagger}$ & \textit{0.406} & \textit{0.227} & \textit{0.334} & \textit{0.349} & \textit{\textbf{0.154}} & \textit{0.240} & \textit{\textbf{0.310}} & \textit{\textbf{0.143}} & \textit{0.230} \\
\rowcolor{gray!12}\osel{}(\sprefine{})$^{\dagger}$ & \textit{0.401} & \textit{0.224} & \textit{0.338} & \textit{0.350} & \textit{0.153} & \textit{0.255} & \textit{0.306} & \textit{0.142} & \textit{0.232} \\
\rowcolor{gray!12}\osel{}(\oregenp{})$^{\dagger}$ & \textit{0.405} & \textit{0.227} & \textit{0.315} & \textit{0.335} & \textit{0.150} & \textit{0.211} & \textit{0.301} & \textit{0.134} & \textit{0.218} \\
\bottomrule
\end{tabular}
\caption{Oracle variants on the full test sets, with \spinfer{} and \sprefine{} for reference. $^{\dagger}$Shaded rows are oracles: they see the reference, directly or through reference-anchored judge scores used for selection, and are not deployable. Best per column within each generator in \textbf{bold}.}
\label{tab:ext}
\end{table*}

\begin{table}[H]
\centering
\scriptsize
\setlength{\tabcolsep}{2.4pt}
\begin{tabular}{@{}l cc cc cc@{}}
\toprule
 & \multicolumn{2}{c}{\textbf{Abstract}} & \multicolumn{2}{c}{\textbf{Review}} & \multicolumn{2}{c}{\textbf{Topic}} \\
\cmidrule(lr){2-3}\cmidrule(lr){4-5}\cmidrule(lr){6-7}
\textbf{Method} & R-L & MET & R-L & MET & R-L & MET \\
\midrule
\multicolumn{7}{@{}l}{\textit{DeepSeek-V4-Flash}} \\
\spinfer{} & 0.269 & 0.356 & 0.167 & 0.273 & 0.147 & 0.241 \\
\spregen{} & 0.283 & 0.368 & 0.163 & 0.283 & 0.149 & 0.245 \\
\rowcolor{gray!12}\oregen{}$^{\dagger}$ & \textbf{0.391} & \textbf{0.474} & \textbf{0.253} & \textbf{0.345} & \textbf{0.217} & \textbf{0.309} \\
\midrule
\multicolumn{7}{@{}l}{\textit{Grok-4.1-Fast}} \\
\spinfer{} & 0.248 & 0.307 & 0.150 & 0.254 & 0.149 & 0.242 \\
\spregen{} & 0.255 & 0.290 & 0.148 & 0.247 & 0.150 & 0.225 \\
\rowcolor{gray!12}\oregen{}$^{\dagger}$ & \textbf{0.381} & \textbf{0.409} & \textbf{0.208} & \textbf{0.318} & \textbf{0.279} & \textbf{0.347} \\
\midrule
\multicolumn{7}{@{}l}{\textit{GPT-5.4-nano}} \\
\spinfer{} & 0.214 & 0.328 & 0.149 & \textbf{0.251} & 0.130 & \textbf{0.220} \\
\spregen{} & 0.223 & 0.324 & 0.148 & 0.236 & 0.128 & 0.208 \\
\rowcolor{gray!12}\oregen{}$^{\dagger}$ & \textbf{0.232} & \textbf{0.331} & \textbf{0.153} & 0.234 & \textbf{0.134} & 0.219 \\
\bottomrule
\end{tabular}
\caption{Oracle revision vs.\ its reference-free control on the same random test instances (498/488/494 for abstract/review/topic). $^{\dagger}$Oracle (sees the reference). Best per column within each generator in \textbf{bold}.}
\label{tab:control}
\end{table}

\section{Per-Criterion LLM-Judge Scores}
\label{app:judge}
Tables~\ref{tab:judge-DeepSeek}--\ref{tab:judge-GPT} give all nine criterion scores (GPT-5.4-nano judge, full test sets; for DeepSeek \sprefine{} on abstracts, 4,557 of the 4,560 instances have an output).

\newcommand{\judgehead}{\toprule
\textbf{Task} & \textbf{Method} & Lex. & Syn. & Tone & Disc. & Fmt. & \sfid{} & Cov. & Fact. & Flu. & Len. & \tq{} \\
\midrule}
\newcommand{\judgecap}[1]{Per-criterion LLM-judge scores (1--5) for #1. Style: Lexical, Syntax, Tone, Discourse, Format; Task Quality: Coverage, Factuality, Fluency, Length. Best per column within each task in \textbf{bold}.}

\begin{table*}[t]
\centering\scriptsize\setlength{\tabcolsep}{4pt}
\begin{tabular}{@{}ll ccccc c cccc c@{}}
\judgehead
\multirow{6}{*}{Abstract} & \noprof{} & 2.62 & 2.74 & 3.61 & 2.81 & 2.24 & 2.80 & 2.88 & 3.31 & 4.54 & 2.95 & 3.42 \\
 & \randuser{} & 2.64 & 2.84 & 3.62 & 2.84 & 3.24 & 3.04 & 2.95 & 3.09 & 4.52 & 2.89 & 3.36 \\
 & \longlamp{} & 2.83 & 2.95 & 3.73 & 2.99 & 3.08 & 3.12 & 3.16 & 3.08 & 4.55 & 2.90 & 3.42 \\
 & \spprofile{} & 2.85 & 3.16 & 3.84 & 3.12 & \textbf{4.31} & 3.46 & 3.01 & \textbf{3.33} & 4.57 & 3.14 & 3.51 \\
 & \spinfer{} & 2.92 & \textbf{3.17} & 3.88 & \textbf{3.19} & 3.83 & 3.40 & 3.13 & 3.32 & 4.60 & \textbf{3.18} & \textbf{3.56} \\
\rowcolor{refinebg} & \sprefine{} & \textbf{2.93} & \textbf{3.17} & \textbf{3.90} & 3.18 & 4.26 & \textbf{3.49} & \textbf{3.23} & 3.23 & \textbf{4.61} & 3.06 & 3.53 \\
\midrule
\multirow{6}{*}{Review} & \noprof{} & 1.56 & 2.05 & 2.53 & 2.09 & 1.79 & 2.00 & 1.88 & \textbf{3.37} & 4.55 & 2.55 & 3.09 \\
 & \randuser{} & 1.67 & 2.17 & 2.69 & 2.22 & 2.43 & 2.24 & 1.94 & 3.32 & \textbf{4.57} & 2.59 & 3.10 \\
 & \longlamp{} & 1.71 & 2.21 & 2.72 & 2.27 & 2.39 & 2.26 & 1.98 & 3.32 & 4.56 & 2.66 & 3.13 \\
 & \spprofile{} & 1.83 & 2.41 & 2.94 & 2.54 & 3.32 & 2.61 & 2.02 & 3.32 & 4.53 & \textbf{2.92} & \textbf{3.20} \\
 & \spinfer{} & \textbf{1.92} & \textbf{2.48} & \textbf{3.01} & 2.63 & 2.31 & 2.47 & 2.20 & 3.09 & 4.52 & 2.90 & 3.18 \\
\rowcolor{refinebg} & \sprefine{} & 1.90 & 2.47 & \textbf{3.01} & \textbf{2.69} & \textbf{3.37} & \textbf{2.69} & \textbf{2.22} & 3.02 & 4.56 & 2.87 & 3.17 \\
\midrule
\multirow{6}{*}{Topic} & \noprof{} & 1.35 & 1.80 & 2.13 & 1.90 & 1.93 & 1.82 & 1.80 & 2.66 & \textbf{4.29} & 2.12 & 2.72 \\
 & \randuser{} & 1.36 & 1.76 & 2.09 & 1.78 & 2.06 & 1.81 & 1.59 & 2.68 & 4.03 & 1.97 & 2.57 \\
 & \longlamp{} & 1.44 & 1.82 & 2.18 & 1.80 & 2.07 & 1.86 & 1.69 & 2.72 & 4.02 & 1.98 & 2.60 \\
 & \spprofile{} & 1.53 & 2.16 & 2.48 & 2.18 & 2.78 & 2.23 & 1.73 & \textbf{3.00} & 4.19 & 2.35 & 2.82 \\
 & \spinfer{} & \textbf{1.62} & \textbf{2.24} & \textbf{2.56} & \textbf{2.33} & \textbf{2.87} & \textbf{2.32} & 1.87 & 2.91 & 4.27 & \textbf{2.41} & \textbf{2.86} \\
\rowcolor{refinebg} & \sprefine{} & 1.59 & 2.21 & 2.48 & 2.30 & 2.81 & 2.28 & \textbf{1.90} & 2.78 & 4.21 & 2.34 & 2.81 \\
\bottomrule
\end{tabular}
\caption{\judgecap{DeepSeek-V4-Flash}}
\label{tab:judge-DeepSeek}
\end{table*}

\begin{table*}[t]
\centering\scriptsize\setlength{\tabcolsep}{4pt}
\begin{tabular}{@{}ll ccccc c cccc c@{}}
\judgehead
\multirow{6}{*}{Abstract} & \noprof{} & 2.54 & 2.66 & 3.46 & 2.78 & 3.53 & 2.99 & 3.14 & 2.13 & 4.48 & 2.70 & 3.11 \\
 & \randuser{} & 2.57 & 2.70 & 3.51 & 2.82 & 3.81 & 3.08 & 3.12 & 2.39 & 4.53 & 2.79 & 3.21 \\
 & \longlamp{} & 2.81 & 2.91 & 3.69 & 3.03 & 3.93 & 3.27 & \textbf{3.31} & 2.63 & \textbf{4.60} & 2.90 & 3.36 \\
 & \spprofile{} & 2.79 & 2.98 & 3.72 & 2.96 & 4.13 & 3.31 & 3.13 & 2.82 & 4.51 & 3.00 & 3.36 \\
 & \spinfer{} & 2.84 & 3.07 & 3.80 & \textbf{3.07} & \textbf{4.22} & \textbf{3.40} & 3.05 & 3.21 & 4.57 & \textbf{3.19} & \textbf{3.51} \\
\rowcolor{refinebg} & \sprefine{} & \textbf{2.91} & \textbf{3.10} & \textbf{3.84} & 3.04 & 4.14 & \textbf{3.40} & 2.99 & \textbf{3.32} & 4.46 & 3.09 & 3.46 \\
\midrule
\multirow{6}{*}{Review} & \noprof{} & 1.46 & 1.89 & 2.21 & 2.02 & 1.40 & 1.80 & 2.08 & 2.43 & \textbf{4.40} & 2.10 & 2.75 \\
 & \randuser{} & 1.48 & 1.88 & 2.11 & 1.99 & 1.57 & 1.81 & 2.05 & 2.41 & 4.33 & 1.94 & 2.68 \\
 & \longlamp{} & 1.70 & 2.09 & 2.51 & 2.27 & 1.77 & 2.07 & \textbf{2.25} & 2.35 & 4.37 & 2.04 & 2.75 \\
 & \spprofile{} & 1.74 & 2.22 & 2.60 & 2.29 & \textbf{2.84} & \textbf{2.34} & 2.04 & \textbf{2.97} & 4.38 & \textbf{2.72} & 3.03 \\
 & \spinfer{} & 1.89 & 2.40 & 2.86 & \textbf{2.48} & 1.71 & 2.27 & 2.18 & 2.84 & \textbf{4.40} & \textbf{2.72} & \textbf{3.04} \\
\rowcolor{refinebg} & \sprefine{} & \textbf{1.94} & \textbf{2.44} & \textbf{2.92} & 2.46 & 1.84 & 2.32 & 2.18 & 2.94 & 4.28 & \textbf{2.72} & 3.03 \\
\midrule
\multirow{6}{*}{Topic} & \noprof{} & 1.30 & 1.72 & 2.02 & 1.83 & 2.05 & 1.78 & 1.83 & 2.19 & 4.12 & 1.92 & 2.52 \\
 & \randuser{} & 1.36 & 1.82 & 2.02 & 1.91 & 2.26 & 1.88 & 1.74 & 2.31 & 4.07 & 2.04 & 2.54 \\
 & \longlamp{} & 1.57 & 2.04 & 2.29 & 2.14 & 2.47 & 2.10 & \textbf{1.96} & 2.42 & \textbf{4.13} & 2.17 & 2.67 \\
 & \spprofile{} & 1.45 & 1.96 & 2.06 & 2.00 & 2.47 & 1.99 & 1.79 & 2.37 & 3.96 & 2.20 & 2.58 \\
 & \spinfer{} & 1.59 & 2.13 & 2.38 & \textbf{2.21} & \textbf{2.64} & \textbf{2.19} & 1.89 & \textbf{2.67} & \textbf{4.13} & \textbf{2.33} & \textbf{2.75} \\
\rowcolor{refinebg} & \sprefine{} & \textbf{1.63} & \textbf{2.19} & \textbf{2.39} & 2.15 & 2.54 & 2.18 & 1.90 & 2.60 & 3.98 & 2.32 & 2.70 \\
\bottomrule
\end{tabular}
\caption{\judgecap{Grok-4.1-Fast}}
\label{tab:judge-Grok}
\end{table*}

\begin{table*}[t]
\centering\scriptsize\setlength{\tabcolsep}{4pt}
\begin{tabular}{@{}ll ccccc c cccc c@{}}
\judgehead
\multirow{6}{*}{Abstract} & \noprof{} & 2.66 & 2.85 & 3.65 & 2.81 & 3.52 & 3.10 & 2.83 & \textbf{3.79} & 4.63 & 3.06 & 3.58 \\
 & \randuser{} & 2.67 & 2.91 & 3.72 & 2.91 & 3.58 & 3.16 & 2.97 & 3.75 & 4.68 & 3.08 & 3.62 \\
 & \longlamp{} & 2.80 & 2.99 & 3.79 & 3.04 & 3.60 & 3.24 & 3.12 & 3.76 & 4.70 & \textbf{3.09} & \textbf{3.67} \\
 & \spprofile{} & 2.77 & \textbf{3.02} & 3.85 & 3.05 & \textbf{4.25} & \textbf{3.39} & 3.07 & 3.73 & 4.64 & 2.93 & 3.59 \\
 & \spinfer{} & 2.78 & 2.98 & 3.85 & 3.10 & 4.07 & 3.36 & \textbf{3.20} & 3.66 & \textbf{4.71} & 2.81 & 3.60 \\
\rowcolor{refinebg} & \sprefine{} & \textbf{2.84} & 3.01 & \textbf{3.88} & \textbf{3.12} & 4.10 & \textbf{3.39} & \textbf{3.20} & 3.69 & 4.68 & 2.79 & 3.59 \\
\midrule
\multirow{6}{*}{Review} & \noprof{} & 1.51 & 2.01 & 2.49 & 2.03 & 2.70 & 2.15 & 1.65 & 3.81 & 4.60 & 2.10 & 3.04 \\
 & \randuser{} & 1.58 & 2.13 & 2.59 & 2.31 & 2.83 & 2.29 & 1.82 & 3.75 & \textbf{4.69} & 2.66 & 3.23 \\
 & \longlamp{} & 1.65 & 2.17 & 2.65 & 2.35 & 2.89 & 2.34 & 1.86 & 3.76 & 4.68 & 2.71 & 3.25 \\
 & \spprofile{} & 1.67 & \textbf{2.23} & 2.72 & \textbf{2.44} & \textbf{3.36} & \textbf{2.48} & 1.86 & \textbf{3.82} & 4.65 & \textbf{2.79} & \textbf{3.28} \\
 & \spinfer{} & \textbf{1.68} & 2.22 & \textbf{2.76} & 2.38 & 1.84 & 2.17 & \textbf{1.95} & 3.67 & 4.61 & 2.75 & 3.24 \\
\rowcolor{refinebg} & \sprefine{} & \textbf{1.68} & \textbf{2.23} & 2.72 & 2.38 & 1.91 & 2.18 & \textbf{1.95} & 3.68 & 4.61 & 2.74 & 3.25 \\
\midrule
\multirow{6}{*}{Topic} & \noprof{} & 1.27 & 1.73 & 2.09 & 1.86 & 2.07 & 1.80 & 1.62 & \textbf{3.24} & 4.39 & 2.04 & 2.82 \\
 & \randuser{} & 1.31 & 1.76 & 2.11 & 1.86 & 2.06 & 1.82 & 1.65 & 3.03 & 4.29 & 1.99 & 2.74 \\
 & \longlamp{} & 1.36 & 1.81 & 2.19 & 1.91 & 2.06 & 1.87 & 1.73 & 3.07 & 4.32 & 2.07 & 2.80 \\
 & \spprofile{} & 1.41 & 2.04 & 2.40 & 2.16 & \textbf{2.69} & \textbf{2.14} & 1.67 & 3.22 & 4.30 & 2.25 & 2.86 \\
 & \spinfer{} & \textbf{1.46} & 2.06 & \textbf{2.45} & \textbf{2.17} & 2.53 & \textbf{2.14} & \textbf{1.79} & \textbf{3.24} & \textbf{4.42} & \textbf{2.31} & \textbf{2.94} \\
\rowcolor{refinebg} & \sprefine{} & \textbf{1.46} & \textbf{2.08} & 2.43 & 2.15 & 2.54 & 2.13 & \textbf{1.79} & 3.23 & 4.38 & 2.27 & 2.92 \\
\bottomrule
\end{tabular}
\caption{\judgecap{GPT-5.4-nano}}
\label{tab:judge-GPT}
\end{table*}

\clearpage
\makeatletter\global\@colht\textheight\global\@colroom\textheight\global\vsize\textheight\makeatother
\section{Prompts and Algorithms}
\label{app:prompts}
Runtime fields are shown in braces. Samples are rendered per task as ``Title: \dots{} Abstract: \dots'' (abstracts), ``Rating: \dots{} Product: \dots{} Summary: \dots{} Review: \dots'' (reviews), or ``Summary: \dots{} Post: \dots'' (topics), and numbered with ``-- SAMPLE $i$ --'' separators.

\subsection{\sprefine{} (all four steps, in full)}
\begin{tcblisting}{promptbox, listing only, breakable, title={\sprefine{} step 1: \textsc{ExtractStyle} ($\phi_{\text{ext}}$; also call 1 of \spprofile{})}, listing options={basicstyle=\ttfamily\scriptsize, breaklines=true, breakatwhitespace=false, columns=fullflexible, keepspaces=true, showstringspaces=false}}
You are an expert computational linguist specializing in writing analysis. Below are {n} writing samples from ONE author. Read all carefully and build a Personal Style Profile (PSP).

Rules:
- Base all claims on the text
- Quote exact phrases as evidence
- Do NOT infer patterns not present
- If not observable, write: "not observable from samples"

=======================
SAMPLES
=======================
{samples}
=======================

Analyze across ALL categories: lexical choice, syntactic patterns, discourse markers, structural organization, pragmatic tone, formatting & layout, cross-task variability.

Return ONLY valid JSON with this schema (replace the "..." values with your analysis):

{
  "lexical_choice": {
    "word_frequency_and_preference": {"pattern": "...", "evidence": ["..."], "avoid": "..."},
    "stopword_avoidance":            {"pattern": "...", "evidence": ["..."], "avoid": "..."},
    "domain_specific_lexicon":       {"pattern": "...", "evidence": ["..."], "avoid": "..."},
    "intensifier_and_modifier_use":  {"pattern": "...", "evidence": ["..."], "avoid": "..."},
    "hedging_behavior":              {"pattern": "...", "evidence": ["..."], "avoid": "..."}
  },
  "syntactic_patterns": {
    "sentence_length_and_complexity":  {"avg_words_per_sentence": 0, "pattern": "...", "evidence": ["..."], "avoid": "..."},
    "passive_vs_active_voice":         {"dominant": "...", "passive_percentage": "...", "pattern": "...", "evidence": ["..."], "avoid": "..."},
    "coordination_vs_subordination":   {"dominant": "...", "pattern": "...", "evidence": ["..."], "avoid": "..."},
    "punctuation_habits":              {"pattern": "...", "evidence": ["..."], "avoid": "..."},
    "enumerative_style":               {"uses_enumeration": false, "format": "...", "evidence": ["..."], "avoid": "..."}
  },
  "discourse_markers": {
    "transitional_phrases": {"top_phrases": ["..."], "evidence": ["..."], "avoid": "..."},
    "opening_moves":        {"pattern": "...", "evidence": ["..."], "avoid": "..."},
    "closing_moves":        {"pattern": "...", "evidence": ["..."], "avoid": "..."},
    "reference_density":    {"frequency": "...", "format": "...", "evidence": ["..."]}
  },
  "structural_organization": {
    "paragraph_structure":         {"avg_sentences_per_paragraph": 0, "style": "...", "evidence": ["..."], "avoid": "..."},
    "rhetorical_role_distribution":{"pattern": "...", "sequence": "...", "evidence": ["..."], "avoid": "..."},
    "argument_framing":            {"pattern": "...", "evidence": ["..."], "avoid": "..."},
    "section_naming_style":        {"pattern": "...", "evidence": ["..."]}
  },
  "pragmatic_tone": {
    "formality_level":         {"level": "...", "pattern": "...", "evidence": ["..."], "avoid": "..."},
    "politeness_strategy":     {"style": "...", "pattern": "...", "evidence": ["..."], "avoid": "..."},
    "audience_awareness":      {"pronoun": "...", "pattern": "...", "evidence": ["..."], "avoid": "..."},
    "figurative_language":     {"uses_figurative_language": false, "pattern": "...", "evidence": ["..."], "avoid": "..."},
    "certainty_and_commitment":{"stance": "...", "strong_verbs": ["..."], "hedge_words": ["..."], "evidence": ["..."], "avoid": "..."}
  },
  "formatting_layout": {
    "visual_structure":           {"uses_bullets": false, "uses_numbered_lists": false, "uses_headers": false, "pattern": "...", "evidence": ["..."]},
    "paragraph_density":          {"style": "...", "evidence": ["..."], "avoid": "..."},
    "emphasis_and_capitalization":{"uses_emphasis": false, "methods": ["none"], "evidence": ["..."]}
  },
  "cross_task_variability": {
    "style_adaptation_by_task":   {"pattern": "...", "evidence": ["..."]},
    "content_style_entanglement": {"fingerprint_patterns": ["..."], "evidence": ["..."], "note": "..."}
  }
}
\end{tcblisting}
\begin{tcblisting}{promptbox, listing only, breakable, title={\sprefine{} step 2: \textsc{Draft} ($\phi_{\text{draft}}$; the entire \spinfer{} method)}, listing options={basicstyle=\ttfamily\scriptsize, breaklines=true, breakatwhitespace=false, columns=fullflexible, keepspaces=true, showstringspaces=false}}
You are writing in the style of a specific author. Below are {k} writing samples from this author. Read them carefully, infer the author's style, and then complete the task in that style. Do not produce a written analysis -- internalize the style and write.

=======================
AUTHOR'S WRITING SAMPLES
=======================
{numbered_samples}

=======================
TASK
=======================
Task type: {task_type}
Task input:
{task_input}

=======================
INSTRUCTIONS
=======================
1. Silently identify the author's recurring style features: sentence rhythm, word choices, hedging, transitions, paragraph structure, tone, formatting.
2. Draft the task output matching those features.
3. Briefly self-check that your draft matches the style of the samples; revise if it drifts.
4. Output ONLY the final text -- no preamble, no analysis, no fences.
\end{tcblisting}
\begin{tcblisting}{promptbox, listing only, breakable, title={\sprefine{} step 3: \textsc{Revise} ($\phi_{\text{rev}}$)}, listing options={basicstyle=\ttfamily\scriptsize, breaklines=true, breakatwhitespace=false, columns=fullflexible, keepspaces=true, showstringspaces=false}}
You are an expert writing evaluator and ghostwriter. You will complete two steps in this call.

=======================
AUTHOR'S WRITING SAMPLES
(k=4 retrieved profile entries -- your style reference for this call)
=======================
{numbered_samples}

=======================
PERSONAL STYLE PROFILE (PSP)
(structured style specification extracted from the author's writing history)
=======================
{psp_json}

=======================
TASK REMINDER
=======================
Task type: {task_type}
Task input:
{task_input}

Your output must complete this task and only this task. Do not change the subject matter.

=======================
INITIAL DRAFT TO IMPROVE
(generated by SP-Infer)
=======================
{initial_draft}

=======================
STEP 1 -- PSP-CONDITIONED REVISION
=======================
Using the PSP and writing samples as style anchors, revise the initial draft to better match the author's style. Apply every applicable PSP field: sentence length, voice, hedging, transitional phrases, paragraph density, and formatting. Do not alter factual content. Do not introduce new facts not present in the task input.

Produce a revised draft. This is your Round 1 draft.

=======================
STEP 2 -- DIMENSION A SCORING
(score your Round 1 draft against the author's writing samples above)
=======================
Score the Round 1 draft on each Dimension A criterion below using a 1-5 scale.

Scoring scale:
  1 = Poor  |  2 = Below average  |  3 = Adequate  |  4 = Good  |  5 = Excellent

DIMENSION A -- STYLISTIC FIDELITY (compare Round 1 draft to the author's writing samples):
  A1. Lexical Match         -- vocabulary is entirely generic with no overlap (1) -> word choices indistinguishable from the author (5)
  A2. Syntactic Match       -- sentence length and voice are opposite to the author (1) -> length, complexity, and voice indistinguishable from the author (5)
  A3. Tone & Formality      -- register is completely wrong (1) -> formality, hedging, and certainty stance indistinguishable from the author (5)
  A4. Discourse & Structure -- no recognizable transitions or paragraph density (1) -> discourse organization indistinguishable from the author (5)
  A5. Formatting            -- formatting is opposite to the author's (1) -> bullets, headers, and prose structure indistinguishable from the author (5)

Return your Dimension A scores in this exact JSON format:
```json
{
  "round": 1,
  "dimension_a_scores": {
    "A1_lexical_match": 0,
    "A2_syntactic_match": 0,
    "A3_tone_formality": 0,
    "A4_discourse_structure": 0,
    "A5_formatting": 0
  },
  "style_fidelity_score": 0.0,
  "weakest_criteria": ["...", "..."],
  "specific_issues": "..."
}
```

=======================
OUTPUT FORMAT
=======================
Return your response in exactly this structure:

ROUND 1 DRAFT:
<your revised draft text only, no preamble, no fences>

DIMENSION A SCORES (ROUND 1):
```json
<your scores JSON from Step 2>
```
\end{tcblisting}
\begin{tcblisting}{promptbox, listing only, breakable, title={\sprefine{} step 4: \textsc{TargetedRevise} ($\phi_{\text{tgt}}$)}, listing options={basicstyle=\ttfamily\scriptsize, breaklines=true, breakatwhitespace=false, columns=fullflexible, keepspaces=true, showstringspaces=false}}
You are an expert writing evaluator and ghostwriter continuing an iterative style refinement pipeline.

=======================
AUTHOR'S WRITING SAMPLES
(k=4 retrieved profile entries -- your style reference for this call)
=======================
{numbered_samples}

=======================
PERSONAL STYLE PROFILE (PSP)
=======================
{psp_json}

=======================
TASK REMINDER
=======================
Task type: {task_type}
Task input:
{task_input}

Your output must complete this task and only this task. Do not change the subject matter.

=======================
ROUND 1 DRAFT
(produced in the previous call)
=======================
{round1_draft}

=======================
ROUND 1 DIMENSION A SCORES
(from the previous call -- use these to guide your revision)
=======================
{round1_scores_json}

=======================
STEP 1 -- TARGETED REVISION
=======================
Using the Round 1 Dimension A scores and specific issues above, revise the Round 1 draft.

Revision priorities:
1. Fix criteria scored 1 or 2 first -- these are critical style failures.
2. Then improve criteria scored 3 -- bring them to 4 or 5.
3. Do not change criteria already scoring 4 or 5.
4. Style corrections must not alter factual content.
5. Do not introduce new facts not present in the task input.

Produce a revised draft. This is your Round 2 draft.

=======================
STEP 2 -- DIMENSION A RESCORE
(score your Round 2 draft against the author's writing samples above)
=======================
Score the Round 2 draft on each Dimension A criterion below using a 1-5 scale.

Scoring scale:
  1 = Poor  |  2 = Below average  |  3 = Adequate  |  4 = Good  |  5 = Excellent

DIMENSION A -- STYLISTIC FIDELITY (compare Round 2 draft to the author's writing samples):
  A1. Lexical Match         -- vocabulary is entirely generic with no overlap (1) -> word choices indistinguishable from the author (5)
  A2. Syntactic Match       -- sentence length and voice are opposite to the author (1) -> length, complexity, and voice indistinguishable from the author (5)
  A3. Tone & Formality      -- register is completely wrong (1) -> formality, hedging, and certainty stance indistinguishable from the author (5)
  A4. Discourse & Structure -- no recognizable transitions or paragraph density (1) -> discourse organization indistinguishable from the author (5)
  A5. Formatting            -- formatting is opposite to the author's (1) -> bullets, headers, and prose structure indistinguishable from the author (5)

Return your Dimension A scores in this exact JSON format:
```json
{
  "round": 2,
  "dimension_a_scores": {
    "A1_lexical_match": 0,
    "A2_syntactic_match": 0,
    "A3_tone_formality": 0,
    "A4_discourse_structure": 0,
    "A5_formatting": 0
  },
  "style_fidelity_score": 0.0,
  "weakest_criteria": ["...", "..."],
  "specific_issues": "..."
}
```

=======================
OUTPUT FORMAT
=======================
Return your response in exactly this structure:

ROUND 2 DRAFT:
<your revised draft text only, no preamble, no fences>

DIMENSION A SCORES (ROUND 2):
```json
<your scores JSON from Step 2>
```
\end{tcblisting}

\subsection{\spprofile{}}
Call 1 is \textsc{ExtractStyle} above; call 2 is:
\begin{tcblisting}{promptbox, listing only, breakable, title={\spprofile{} call 2: \textsc{Apply} ($\phi_{\text{apply}}$)}, listing options={basicstyle=\ttfamily\scriptsize, breaklines=true, breakatwhitespace=false, columns=fullflexible, keepspaces=true, showstringspaces=false}}
You are writing in the style of a specific author. Below is their Personal Style Profile (PSP), built from analyzing prior writing samples. Your job is to produce text that matches this PSP as closely as possible while completing the requested task accurately.

=======================
PERSONAL STYLE PROFILE
=======================
{psp_json}

=======================
TASK
=======================
Task type: {task_type}
Task input:
{task_input}

=======================
INSTRUCTIONS
=======================
Step 1 -- DRAFT (silent): Write a draft that completes the task. While drafting, actively conform to every applicable PSP field -- sentence length, voice, hedging, transitional phrases, paragraph density, formatting choices. Use the "evidence" phrases as stylistic anchors but do NOT copy them verbatim if they would be content-irrelevant to this task.

Step 2 -- REVISE (silent): Re-read your draft against the PSP. For each PSP category, ask: does the draft match the stated pattern? If a PSP field says "avoid X," confirm X is absent. If avg_words_per_sentence is N, check your draft is within +/-20% of N. If passive_percentage is "low," count passives. Rewrite anywhere the draft drifts.

Step 3 -- RESPECT TASK CONSTRAINTS: The output must still be a faithful completion of the task input. Style is a constraint on HOW you write, not WHAT you write. Do not invent facts. Do not add bullets/headers unless the PSP indicates the author uses them.

=======================
OUTPUT
=======================
Return ONLY the final revised text. No explanation, no preamble, no markdown fences, no "Here is the abstract:" -- just the text itself.
\end{tcblisting}

\subsection{Baselines}
\begin{tcblisting}{promptbox, listing only, breakable, title={\longlamp{} (LongLaMP's templates; $e_1,\dots,e_k$ are the retrieved entries)}, listing options={basicstyle=\ttfamily\scriptsize, breaklines=true, breakatwhitespace=false, columns=fullflexible, keepspaces=true, showstringspaces=false}}
Abstract:  "{abstract of e_1}" is the abstract for the title "{title of e_1}", and ... , and "{abstract of e_k}" is the abstract for the title "{title of e_k}". Use the above abstracts as context to understand the style and language of the user and, {task_input}
Review:    "{rating}" is a rating for the product with description "{description}". "{summary}" is summary for "{review text}" , and ... . Following the given patterns {task_input}
Topic:     "{summary}" is a summary for "{post}" , and ... . Following the given patterns, {task_input}
\end{tcblisting}
\begin{tcblisting}{promptbox, listing only, breakable, title={\noprof{} and \randuser{}}, listing options={basicstyle=\ttfamily\scriptsize, breaklines=true, breakatwhitespace=false, columns=fullflexible, keepspaces=true, showstringspaces=false}}
No Profile: the task input alone, e.g.
  Generate an abstract for the title "{title}" using the following items: {items}.

Random-Samples: the LongLaMP template above, filled with k entries sampled uniformly at random
  from the user's own history (not the k entries retrieved by Contriever).
\end{tcblisting}

\subsection{Oracles and controls (abridged)}
\begin{tcblisting}{promptbox, listing only, breakable, title={\oregen{} and \oregenp{} (oracle; abridged)}, listing options={basicstyle=\ttfamily\scriptsize, breaklines=true, breakatwhitespace=false, columns=fullflexible, keepspaces=true, showstringspaces=false}}
You are an expert writing evaluator and ghostwriter. Your task has two steps:
STEP 1 -- evaluate a generated text using the ground truth as a scoring reference only.
STEP 2 -- produce an improved version WITHOUT referencing the ground truth.
[AUTHOR'S WRITING SAMPLES]  [TASK INPUT]  [GENERATED OUTPUT TO EVALUATE]  (SP-Infer output for Oracle-Regen; SP-Profile output for Oracle-Regen+P)
STEP 1 -- EVALUATE (the ground truth below is for scoring reference ONLY -- do not copy it)
Ground truth output (scoring reference only): {ground_truth}
Score 1-5: A1-A5 (compare to the author's writing samples); B1 Content Coverage (relative to ground truth), B2 Factual Consistency, B3 Fluency & Coherence, B4 Length Appropriateness (relative to ground truth). Return JSON with scores, weakest_criteria, specific_issues.
TASK REMINDER: re-read the original task. Do not introduce content from the ground truth.
STEP 2 -- REGENERATE (work only from the task input and author samples -- do NOT copy from the ground truth):
Fix criteria scored 1-2 first, then 3; do not change criteria scoring 4-5; do not alter or add facts. The improved output must: complete the same task using only the task input as content source; match the author's style; be comparable in length to the ground truth; contain no preamble; NOT reproduce phrases or sentences from the ground truth.
Return: EVALUATION: {json}  REGENERATED OUTPUT: <text>
\end{tcblisting}

\begin{tcblisting}{promptbox, listing only, breakable, title={\spregen{} and \spregenp{} (reference-free controls): changes relative to the oracle prompt}, listing options={basicstyle=\ttfamily\scriptsize, breaklines=true, breakatwhitespace=false, columns=fullflexible, keepspaces=true, showstringspaces=false}}
- Removed: the ground-truth block and every mention of the ground truth.
- B1 Content Coverage: "key information from the task input present and well-represented" (was: relative to ground truth).
- B4 Length Appropriateness: "output length relative to the author's writing samples" (was: relative to ground truth).
- Removed: "Be comparable in length to the ground truth" and "NOT reproduce phrases or sentences from the ground truth".
- All other text identical; same model, same initial drafts.
\end{tcblisting}

\begin{tcblisting}{promptbox, listing only, breakable, title={\osel{} (oracle selection rule)}, listing options={basicstyle=\ttfamily\scriptsize, breaklines=true, breakatwhitespace=false, columns=fullflexible, keepspaces=true, showstringspaces=false}}
For each test instance, score every candidate draft with the LLM judge (reference-anchored) and keep the draft with the highest 0.5 * StyleFidelity + 0.5 * TaskQuality.
Candidate sets: {SP-Infer, Oracle-Regen}; {SP-Refine y_0 (= SP-Infer), y_1, y_2}; {SP-Profile, Oracle-Regen+P}.
\end{tcblisting}

\begin{tcblisting}{promptbox, listing only, breakable, title={LLM judge (abridged)}, listing options={basicstyle=\ttfamily\scriptsize, breaklines=true, breakatwhitespace=false, columns=fullflexible, keepspaces=true, showstringspaces=false}}
You are an expert writing evaluator performing a final blind evaluation of a generated text output.

TASK REMINDER -- Task type: {task_type}  Task input: {task_input}
GENERATED OUTPUT (the text to evaluate): {generated_output}
GROUND TRUTH (what the author actually wrote for this task -- evaluation anchor for both dimensions): {ground_truth}

Score the generated output on all criteria below using a 1-5 scale (1 = Poor, 3 = Adequate, 5 = Excellent). Both dimensions use the ground truth as the evaluation anchor.

DIMENSION A -- STYLISTIC FIDELITY (compare to the ground truth as style reference):
  A1. Lexical Match         -- entirely generic vocabulary (1) -> word choices indistinguishable from ground truth (5)
  A2. Syntactic Match       -- sentence length and voice opposite (1) -> indistinguishable (5)
  A3. Tone & Formality      -- register completely wrong (1) -> formality, hedging, certainty indistinguishable (5)
  A4. Discourse & Structure -- no recognizable transitions or paragraph density (1) -> indistinguishable (5)
  A5. Formatting            -- formatting opposite (1) -> bullets, headers, prose structure indistinguishable (5)

DIMENSION B -- TASK QUALITY (compare to the ground truth):
  B1. Content Coverage       -- most key information missing (1) -> all key information present (5)
  B2. Factual Consistency    -- multiple invented or contradicting facts (1) -> every claim grounded in the task input (5)
  B3. Fluency & Coherence    -- incoherent (1) -> fully fluent and well-structured (5)
  B4. Length Appropriateness -- deviates from ground-truth length by more than 50% (1) -> within ~10% (5)

Return ONLY a JSON object with dimension_a (A1-A5), dimension_b (B1-B4), weakest_criteria, and evaluation_notes.
\end{tcblisting}
\sfid{} and \tq{} are computed by us as the means of the criterion scores; the judge's own aggregate fields are ignored.

\subsection{Notation}
Table~\ref{tab:notation} lists every symbol used in Algorithms~\ref{alg:refine}--\ref{alg:oracle}.

\begin{table}[H]
\centering
\scriptsize
\setlength{\tabcolsep}{3pt}
\begin{tabular}{@{}>{\raggedright\arraybackslash}p{0.33\linewidth} p{0.63\linewidth}@{}}
\toprule
\textbf{Symbol} & \textbf{Meaning} \\
\midrule
$u$, $P_u$ & a user and their history of previously written documents $\{d_1,\dots,d_m\}$ \\
$x$, $c$ & task input (e.g., title and key items) and task type (abstract, review, or topic) \\
$y$, $\hat y$ & the user's reference output (used only for evaluation and by the oracles) and the generated output \\
$R$, $k$ & retriever and number of samples: $R(x, P_u, k)$ returns the $k$ entries of $P_u$ with the highest Contriever dot-product score for $x$ (\S\ref{sec:method}); $k=4$ \\
$S$ & the retrieved samples, $S = R(x, P_u, k)$, shared by all methods \\
$M$, $M(\cdot)$ & the generator LLM; $M(q)$ is one call with prompt $q$ \\
$\Sigma$, $\sigma$ & the \styleprism{} schema of 28 aspects; one aspect \\
$\pi$ & the style profile: for each $\sigma \in \Sigma$, a pattern, quoted evidence $E_\sigma$, and (for 22 aspects) an avoid field \\
$\phi_{\text{ext}}, \phi_{\text{draft}}, \phi_{\text{rev}}, \phi_{\text{tgt}}$ & prompt templates of \sprefine{}'s four steps (\S\ref{app:prompts}) \\
$\phi_{\text{apply}}$, $\phi_{\text{LL}}$, $\phi_{\text{regen}}$ & prompt templates of \spprofile{}'s second call, of \longlamp{}, and of the oracles \\
$y_0, y_1, y_2$ & \sprefine{}'s draft (step 2), revised draft (step 3), and output (step 4); $\hat y = y_2$ \\
$A$, $a_t$ & the five style criteria (Lexical, Syntactic, Tone \& Formality, Discourse \& Structure, Formatting) and $M$'s 1--5 scores of $y_t$ on them, judged against $S$ \\
$s$, $\tilde y$ & an oracle's nine criterion scores and its revised output \\
$\mathcal{C}$ & a set of candidate outputs for \osel{} \\
$\sfid{}(z,y)$, $\tq{}(z,y)$ & the LLM judge's Style Fidelity and Task Quality of candidate $z$, using $y$ as anchor \\
\bottomrule
\end{tabular}
\caption{Notation.}
\label{tab:notation}
\end{table}

\subsection{Algorithms for the ablations and oracles}

\begin{algorithm}[H]
\caption{\spinfer{}: a single call with implicit style inference and self-check.}
\label{alg:infer}
\footnotesize
\begin{algorithmic}[1]
\Require $x, c, P_u, R, k, M$
\State $S \gets R(x, P_u, k)$
\State \Return $\hat y \gets M(\phi_{\text{draft}}(S, c, x))$ \Comment{infer style, draft, self-check}
\end{algorithmic}
\end{algorithm}

\begin{algorithm}[H]
\caption{\spprofile{}: explicit extraction, then generation from the profile without the samples.}
\label{alg:profile}
\footnotesize
\begin{algorithmic}[1]
\Require $x, c, P_u, R, k, M, \Sigma$
\State $S \gets R(x, P_u, k)$
\State $\pi \gets \textsc{ExtractStyle}(S, \Sigma)$ \Comment{as in Alg.~\ref{alg:refine}}
\State \Return $\hat y \gets M(\phi_{\text{apply}}(\pi, c, x))$ \Comment{$S$ not shown}
\end{algorithmic}
\end{algorithm}

\begin{algorithm}[H]
\caption{\oregen{} and its reference-free control \spregen{}; \osel{}. For the +P variants, $y_0$ is the \spprofile{} output instead. $s$ is the reviser's nine criterion scores; $\mathcal{C}$ is a set of candidate outputs.}
\label{alg:oracle}
\footnotesize
\begin{algorithmic}[1]
\Require $x, c, P_u, R, k, M$; reference $y$ (oracles only)
\State $S \gets R(x, P_u, k)$; \ $y_0 \gets$ \spinfer{} output
\State $(s, \tilde y) \gets M(\phi_{\text{regen}}(S, c, x, y_0, y))$ \Comment{\spregen{}: $y$ removed}
\State \Return $\tilde y$
\Statex \textbf{\osel{}:} \Return $\arg\max_{z \in \mathcal{C}} \tfrac12 \sfid{}(z, y) + \tfrac12 \tq{}(z, y)$
\end{algorithmic}
\end{algorithm}

\section{Example Style Profile}
\label{app:psp}
The complete profile $\pi$ extracted for the abstract instance in Figure~\ref{fig:gt-abstract}.
\begin{tcblisting}{promptbox, listing only, breakable, title={Full style profile $\pi$ for idx\_13 (abstract generation, Grok-4.1-Fast), exactly as returned}, listing options={basicstyle=\ttfamily\scriptsize, breaklines=true, breakatwhitespace=false, columns=fullflexible, keepspaces=true, showstringspaces=false}}
{
 "lexical_choice": {
  "word_frequency_and_preference": {
   "pattern": "prefers precise, technical terms with repetition of key concepts",
   "evidence": [
    "tunnelled (repeated 4x in Sample 1)",
    "class B traffic (repeated 6x in Sample 2)",
    "monitoring-cycles (m-cycles) (repeated across Samples 3-4)",
    "fault detection (repeated in Samples 3-4)"
   ],
   "avoid": "not observable from samples"
  },
  "stopword_avoidance": {
   "pattern": "standard use of stopwords, no notable avoidance",
   "evidence": [
    "is an essential element",
    "are wide in scope",
    "are tunnelled over HTTP"
   ],
   "avoid": "not observable from samples"
  },
  "domain_specific_lexicon": {
   "pattern": "heavy use of networking and optical network terminology",
   "evidence": [
    "HTTP tunnels",
    "NetFlow v5",
    "Resilient Packet Ring (RPR)",
    "monitoring-cycles (m-cycles)",
    "all-optical networks (AONs)",
    "NSFNET, ARPA2, SmallNet, and Bellcore"
   ],
   "avoid": "not observable from samples"
  },
  "intensifier_and_modifier_use": {
   "pattern": "sparse use of intensifiers, favors precise qualifiers like 'novel', 'accurate'",
   "evidence": [
    "novel approach",
    "accurate classification",
    "accurate and effective",
    "nearly optimal performance"
   ],
   "avoid": "not observable from samples"
  },
  "hedging_behavior": {
   "pattern": "minimal hedging, uses qualifiers like 'nearly'",
   "evidence": [
    "nearly optimal performance (Samples 3-4)"
   ],
   "avoid": "extensive hedges like 'may', 'possibly'"
  }
 },
 "syntactic_patterns": {
  "sentence_length_and_complexity": {
   "avg_words_per_sentence": 24,
   "pattern": "long, complex sentences with multiple clauses",
   "evidence": [
    "Accurate traffic classification is an essential element of emergent cloud and datacenter architectures. (12 words)",
    "Applications tunnelled over HTTP are wide in scope and diverse in nature, and include mapping, email, video, image, audio and file. (19 words)"
   ],
   "avoid": "not observable from samples"
  },
  "passive_vs_active_voice": {
   "dominant": "mixed, slight passive preference",
   "passive_percentage": "approx 40%",
   "pattern": "passive for technical descriptions",
   "evidence": [
    "is carried out using information",
    "is tested on live data traffic",
    "is formulated as a cycle cover problem"
   ],
   "avoid": "not observable from samples"
  },
  "coordination_vs_subordination": {
   "dominant": "subordination",
   "pattern": "heavy use of subordinate clauses and conjunctions",
   "evidence": [
    "thereby making accurate classification a challenge",
    "using information that is only available from flow-based protocols",
    "competing with background class C traffic"
   ],
   "avoid": "not observable from samples"
  },
  "punctuation_habits": {
   "pattern": "standard use, colons for lists, commas for series",
   "evidence": [
    "namely video, audio and file-transfer.",
    "NSFNET, ARPA2, SmallNet, and Bellcore.",
    "the grade of fault localization, wavelength overhead, and the number of cycles"
   ],
   "avoid": "excessive dashes or semicolons"
  },
  "enumerative_style": {
   "uses_enumeration": true,
   "format": "colon-introduced series or lists",
   "evidence": [
    "namely video, audio and file-transfer.",
    "NSFNET, ARPA2, SmallNet, and Bellcore.",
    "grade of fault localization, wavelength overhead, and the number of cycles"
   ],
   "avoid": "not observable from samples"
  }
 },
 "discourse_markers": {
  "transitional_phrases": {
   "top_phrases": [
    "This paper presents",
    "In this paper",
    "The results show",
    "In comparison to"
   ],
   "evidence": [
    "This paper presents a novel approach (Sample 1)",
    "This paper reports on simulation results (Sample 2)",
    "In this paper, an m-cycle construction (Sample 3)"
   ],
   "avoid": "not observable from samples"
  },
  "opening_moves": {
   "pattern": "problem statement then proposed solution",
   "evidence": [
    "Accurate traffic classification is an essential element... making accurate classification a challenge. This paper presents (Sample 1)",
    "The IEEE P802.17 workgroup is currently standardizing... This paper reports (Sample 2)"
   ],
   "avoid": "not observable from samples"
  },
  "closing_moves": {
   "pattern": "results summary with performance claims",
   "evidence": [
    "the proposed scheme outperforms them by at least 10% (Sample 1)",
    "it achieves nearly optimal performance (Samples 3-4)"
   ],
   "avoid": "not observable from samples"
  },
  "reference_density": {
   "frequency": "low",
   "format": "bracketed numbers",
   "evidence": [
    "[8] (Sample 4)"
   ],
   "avoid": "not observable from samples"
  }
 },
 "structural_organization": {
  "paragraph_structure": {
   "avg_sentences_per_paragraph": 1,
   "style": "single dense paragraph abstracts",
   "evidence": [
    "All samples are single-paragraph abstracts"
   ],
   "avoid": "not observable from samples"
  },
  "rhetorical_role_distribution": {
   "pattern": "problem -> method -> evaluation -> results",
   "sequence": "challenge/problem, proposal, testing/application, performance metrics",
   "evidence": [
    "challenge... This paper presents... tested on... accurately classifies (Sample 1)"
   ],
   "avoid": "not observable from samples"
  },
  "argument_framing": {
   "pattern": "proposal superiority via comparison/metrics",
   "evidence": [
    "outperforms them by at least 10% (Sample 1)",
    "achieves nearly optimal performance (Samples 3-4)"
   ],
   "avoid": "not observable from samples"
  },
  "section_naming_style": {
   "pattern": "descriptive titles with method/network focus",
   "evidence": [
    "Classification of applications in HTTP tunnels.",
    "Spanning-tree based monitoring-cycle construction for fault detection and localization in mesh AONs."
   ]
  }
 },
 "pragmatic_tone": {
  "formality_level": {
   "level": "high",
   "pattern": "academic technical prose",
   "evidence": [
    "emergent cloud and datacenter architectures",
    "formulated as a cycle cover problem with certain constraints"
   ],
   "avoid": "colloquialisms"
  },
  "politeness_strategy": {
   "style": "neutral objective",
   "pattern": "no direct politeness",
   "evidence": "not observable from samples",
   "avoid": "not observable from samples"
  },
  "audience_awareness": {
   "pronoun": "none (impersonal)",
   "pattern": "no personal pronouns",
   "evidence": [
    "We previously showed (Samples 3-4, rare 1st person plural)"
   ],
   "avoid": "1st/2nd person extensively"
  },
  "figurative_language": {
   "uses_figurative_language": false,
   "pattern": "literal only",
   "evidence": "not observable from samples",
   "avoid": "not observable from samples"
  },
  "certainty_and_commitment": {
   "stance": "confident with qualifiers",
   "strong_verbs": [
    "presents",
    "proposes",
    "show",
    "outperforms"
   ],
   "hedge_words": [
    "nearly",
    "at least"
   ],
   "evidence": [
    "accurately classifies at least 70%",
    "achieves nearly optimal performance"
   ],
   "avoid": "not observable from samples"
  }
 },
 "formatting_layout": {
  "visual_structure": {
   "uses_bullets": false,
   "uses_numbered_lists": false,
   "uses_headers": false,
   "pattern": "plain text with -- delimiters in samples",
   "evidence": [
    "-- SAMPLE 1 --",
    "Title: ... Abstract: ..."
   ]
  },
  "paragraph_density": {
   "style": "high density, single block paragraphs",
   "evidence": [
    "All abstracts are unbroken paragraphs"
   ],
   "avoid": "not observable from samples"
  },
  "emphasis_and_capitalization": {
   "uses_emphasis": false,
   "methods": [
    "none"
   ],
   "evidence": [
    "standard title case, no italics/bold"
   ]
  }
 },
 "cross_task_variability": {
  "style_adaptation_by_task": {
   "pattern": "highly consistent across topics",
   "evidence": [
    "Formulaic structure and phrasing repeated: 'This/In this paper... proposed... applied to... results show...' in all samples"
   ]
  },
  "content_style_entanglement": {
   "fingerprint_patterns": [
    "novel/heuristic proposal",
    "network names listed",
    "performance metrics at end",
    "'This/In this paper' opener"
   ],
   "evidence": [
    "novel approach (1), heuristic... proposed (3-4)",
    "NSFNET, ARPA2... (3-4)",
    "accurately classifies at least 70% (1), nearly optimal (3-4)"
   ],
   "note": "Samples 3-4 nearly identical, reinforcing stable style"
  }
 }
}
\end{tcblisting}

\end{document}